\documentclass[lettersize,journal]{IEEEtran}
\usepackage{amsmath,amsfonts}
\usepackage{algorithmic}
\usepackage{algorithm}
\usepackage{array}
\usepackage[caption=false,font=small]{subfig}
\usepackage{textcomp}
\usepackage{stfloats}
\usepackage{url}
\usepackage{verbatim}
\usepackage{graphicx}
\usepackage{cite}

\usepackage[
  colorlinks=true, 
  citecolor=blue,  
  linkcolor=black, 
  urlcolor=black,  
  frenchlinks=true,
  pdfborder={0 0 0}
]{hyperref}
\usepackage{orcidlink}

\usepackage[inkscapelatex=false]{svg}
\usepackage{booktabs}
\usepackage{csquotes}
\include{bib_short.def}
\include{bib_ieee_abbr.def}

\def\BibTeX{{\rm B\kern-.05em{\sc i\kern-.025em b}\kern-.08em
    T\kern-.1667em\lower.7ex\hbox{E}\kern-.125emX}}
\usepackage{balance}

\begin{document}

\title{Context-based Motion Retrieval using Open
Vocabulary Methods for Autonomous Driving
}

\author{Stefan Englmeier\orcidlink{0009-0009-1997-989X}, Max A. Büttner\orcidlink{0009-0006-7939-6256}, Katharina Winter\orcidlink{0009-0002-1952-6558}, and Fabian B. Flohr\orcidlink{0000-0002-1499-3790}

\thanks{
This work received joint funding from the German Federal Minister for Economic Affairs and Energy (BMWE) within the projects STADT:up and NXT GEN AI METHODS (grant no. 19A22006N and 19A23914M) and the Federal Ministry of Research, Technology and Space (BMFTR) within the project ADRIVE-GPT (grant no. 13FH544KA2). The authors are solely responsible for the content of this publication.}

\thanks{The authors are with the Munich University of Applied Sciences, Intelligent Vehicles Lab (IVL), 80335 Munich, Germany (e-mail: intelligent-vehicles@hm.edu)}}
\maketitle

\begin{abstract}
Autonomous driving systems must operate reliably in safety-critical scenarios, particularly those involving unusual or complex behavior by Vulnerable Road Users (VRUs). Identifying these edge cases in driving datasets is essential for robust evaluation and generalization, but retrieving such rare human behavior scenarios within the long tail of large-scale datasets is challenging. To support targeted evaluation of autonomous driving systems in diverse, human-centered scenarios, we propose a novel context-aware motion retrieval framework. Our method combines Skinned Multi-Person Linear (SMPL)-based motion sequences and corresponding video frames before encoding them into a shared multimodal embedding space aligned with natural language. Our approach enables the scalable retrieval of human behavior and their context through text queries.
This work also introduces our dataset WayMoCo, an extension of the Waymo Open Dataset. It contains automatically labeled motion and scene context descriptions derived from generated pseudo-ground-truth SMPL sequences and corresponding image data. Our approach outperforms state-of-the-art models by up to 27.5\% accuracy in motion-context retrieval, when evaluated on the WayMoCo dataset. 

\end{abstract}


\section{Introduction}
Training and evaluating machine learning models for autonomous driving heavily depends on the availability of large-scale, high-quality data. Datasets such as the Waymo Open Dataset (WOD) \cite{Sun_2020_CVPR} have been collected for this purpose, providing vast collections of real-world driving scenarios. However, effective evaluation requires identifying rare, complex and safety-critical situations, which are generally found in the long-tail distribution of these datasets, making efficient utilization of large volumes of data a major challenge in research and development. 
Uniform sampling is unlikely to capture such corner cases appropriately, while manual sampling is tedious, rendering these approaches unsuitable for meaningful evaluation.
Existing work has already explored retrieval‑based methods to pinpoint relevant video segments in autonomous driving datasets \cite{zhou2024vision}. 
Interaction scenarios that include Vulnerable Road Users (VRUs) are specifically critical and complex, making the evaluation of a large number of diverse human behaviors indispensable to ensure safe deployment of autonomous driving systems.
Video-retrieval models like TC-CLIP \cite{kim2024tcclip} are capable of filtering specific scenarios by analyzing the full scene, but lack the ability to identify or differentiate the motions of individual people. The Video-Motion model LAVIMO \cite{yin2024tri} fuses Skinned Multi-Person Linear 
(SMPL)~\cite{SMPL:2015} motions with video inputs to identify human motions, but relies on synthetic renderings of human avatars during training, lacking the ability to process real-world videos or capture contextual scene information.
Thus, current approaches are not particularly effective at motion-based scene retrieval from real-world datasets.

\begin{figure}[t]
\centerline{
\includegraphics[width=1.0\columnwidth]{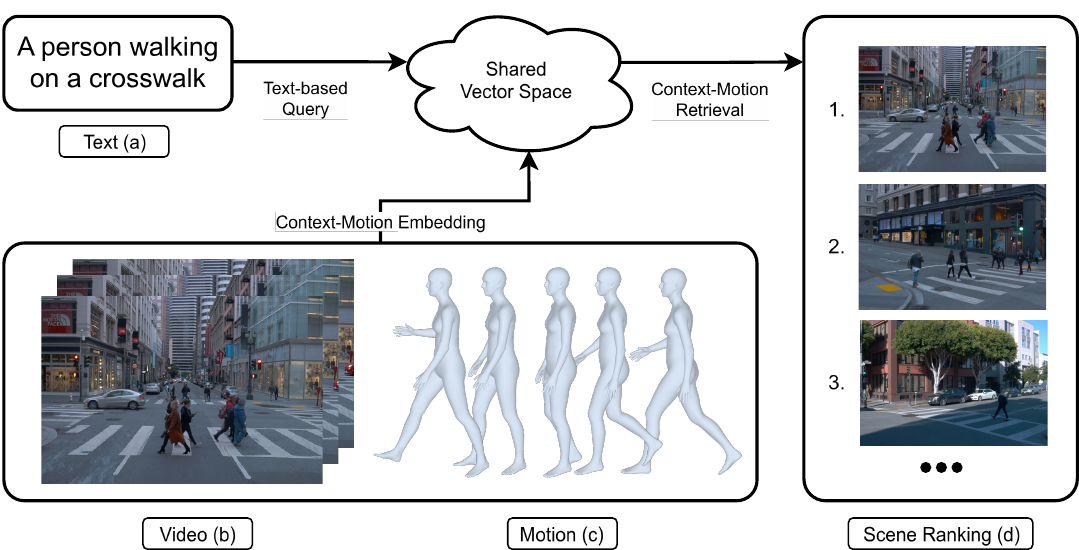}
}
\caption{Overview of our framework: Video (b) and motion (c) sequences are fused into a shared context–motion embedding, while text queries (a) are encoded into the same vector space. This joint representation enables similarity-based retrieval, resulting in a ranked list of matching context-aware motion scenes (d).}
    \label{fig:TitleFigure}
\end{figure}

Our proposed method aims to address this gap by enabling the efficient filtering and selection of rare scenarios with human interaction from large volumes of data (see Figure \ref{fig:TitleFigure}). This is not only crucial for effective evaluation of autonomous driving systems, but can be applied in other research domains, such as anomaly detection in public spaces or assistant robotics.
With natural language queries on human motions, our method allows for filtering of the targeted interaction types and context. 

Our contributions are threefold.
To the best of our knowledge, this is the first context–motion retrieval method to unify SMPL-derived human motion features and real world visual representations from a video encoder into a shared multimodal embedding space. This enables natural language queries to effectively retrieve scenes based on both motion and context.
Second, we present the Waymo Motion Context (WayMoCo) dataset, an extension of the WOD images with natural-language labels for human motions and their contextual scenes. This provides a resource for training and evaluating motion-context retrieval systems.
Third, we conduct a thorough evaluation and ablation study on our newly extended dataset. Our method surpasses state-of-the-art approaches, achieving up to a 27.5\% improvement in motion-context retrieval accuracy over TC-CLIP~\cite{kim2024tcclip}, highlighting the effectiveness of our approach.

We release the full pipeline—including source code, the annotated dataset, and an interactive demo tool\footnote{\url{https://iv.ee.hm.edu/contextmotionclip/}}, allowing researchers to explore, replicate, and apply our method to other datasets with compatible motion and video data.
\section{Related Work}
In this section, we give an overview of state-of-the-art text-based multimodal retrieval methods, highlight different modality fusion mechanisms and present existing human motion datasets. 

\subsection{Multimodal Retrieval}

Contrastive Language–Image Pretraining (CLIP)~\cite{radford2021learning} introduces a powerful joint embedding space \cite{zhang2024vision} for natural language and vision by aligning image and text features through contrastive learning. This open-vocabulary approach enables zero-shot generalization to downstream tasks and forms the foundation for many retrieval and classification models. 

In the video domain, methods such as VideoCLIP \cite{xu-etal-2021-videoclip} and CLIP4Clip~\cite{luo2022clip4clip} extend CLIP’s contrastive training to entire clips. Later works like ViFi-CLIP \cite{hanoonavificlip} and TC-CLIP \cite{kim2024tcclip} further enhance video understanding through temporal modeling and fine-tuned video encoders, achieving strong results in action recognition and retrieval.
In parallel, a line of research applies contrastive alignment to 3D human motions. MotionCLIP \cite{tevet2022motionclip} is among the earliest to map SMPL sequences and captions into CLIP’s joint embedding space, supporting zero-shot classification and generation. Retrieval-focused variants like TMR \cite{petrovich2023tmr} apply a dedicated contrastive loss on a text-to-motion backbone. TMR++ \cite{bensabath2024cross} improve generalization via cross-corpus regularization and LLM-augmented captions. LAVIMO \cite{yin2024tri} adds video as a third input, though it remains confined to synthetic renders, which limits applicability to real-world scenarios.

\subsection{CLIP-Based Multimodal Fusion}

A growing line of work extends CLIP beyond paired image-text data to richer modality sets. AudioCLIP~\cite{guzhov2022audioclip} aligns audio with CLIP's visual-textual space, enabling cross-modal retrieval across all three modalities. CLIP2Point~\cite{huang2023clip2point} maps depth information from point clouds into a CLIP-compatible representation for 3D understanding. M2HF~\cite{journals/corr/abs-2208-07664} facilitates text-video retrieval through a hybrid fusion strategy: visual features extracted from CLIP are early fused with audio and object-motion cues from videos, and subsequently late fused with both the visual and textual representations. 
Across such efforts, lightweight fusion operators—most for example bilinear pooling~\cite{lin2023clip} and attention-based feature mixing~\cite{Chen_Zhong_He_Peng_Zhou_Cheng_2024}—combine modality-specific encoders before contrastive alignment with text. With our model ContextMotionCLIP, follow this fusion paradigm but target a novel combination of SMPL-based human motion and real-world driving video.

\subsection{Human Motion Datasets}
Several large-scale human motion datasets provide SMPL-compatible annotations. AMASS~\cite{AMASS:ICCV:2019} consolidates 15 mocap datasets into the unified SMPL format. Language annotations for the SMPL motions are added using BABEL~\cite{BABEL:CVPR:2021} and HumanML3D~\cite{humanml3d}. EMDB~\cite{kaufmann2023emdb} augments their in-the-wild video with electromagnetic sensors, producing accurate global SMPL fits. Furthermore, recent synthetic datasets target realism at a large scale. BEDLAM~\cite{black2023bedlam} compromises highly accurate renders of 10 subjects per scene, while AGORA~\cite{Agora:CVPR:2021} offers 17k outdoor images, each featuring  between 5 and 15 individuals, both synthetic and annotated with SMPL bodies and detailed 3D fits.

Autonomous driving datasets like Argoverse~\cite{Argoverser:CVPR:2019}, nuScenes~\cite{nuscenes2019}, and Waymo~\cite{Sun_2020_CVPR} include dense 2D/3D joint labels for pedestrians and cyclists — the most structurally similar form of supervision to SMPL. Waymo remains the closest in scale and detail, however, no autonomous driving dataset provides ground-truth SMPL annotations or detailed labels of human motion and context.

\section{Dataset Generation}
\label{method:dataset}
As no existing dataset provides annotated SMPL parameters alongside video data for autonomous driving, we introduce WayMoCo, an extension of the Waymo Open Dataset. All text annotations are automatically labeled with two key types of information: a person’s motion and their context. 
Hence, we create \textit{valid} sequences for our training and evaluation: a 2-second (20-frame) segment of individual VRU's, for which SMPL motion parameters are estimated and paired with automatically generated motion and context labels. The final WayMoCo dataset contains a total of 26,000 such valid sequences.

\subsection{Motion}
\label{sec:data-motion}
To extract motions in the SMPL format from the WOD, we leverage ground-truth bounding boxes and their associated track IDs to construct two-second motion sequences from individual video frames. Based on these bounding boxes, we employ TokenHMR~\cite{dwivedi2024tokenhmr} to estimate human poses for each tracked individual. In order to achieve reliable SMPL extraction, we enforce a minimum bounding box size of $90{\times}35$ pixels. We follow the aforementioned criteria and extract the maximum number of valid motion segments from the dataset: We filter out any leading or trailing frames with a bounding box sized below the minimum threshold. In case more than two frames in a sequence are missing, the sequence is split, if the minimum sequence length is maintained, or discarded otherwise. This way, we achieve non-overlapping, valid sequences from the WOD. 
SMPL root orientation estimates are sometimes unstable under occlusion, resulting in outliers missorientated poses. To address this, we apply Locally Weighted Scatterplot Smoothing (LOWESS)~\cite{cleveland1979robust} to the root orientation sequences, resulting in continuous motion.

For labelling of our valid sequences that contain motion-descriptive terms, we utilize MotionCLIP~\cite{tevet2022motionclip}. To accommodate WOD's 10 Hz frame rate we retrain MotionCLIP.
We use the BABEL-60 motion vocabulary to annotate each sequence, filtering out terms irrelevant to autonomous driving like "play instrument" and removing "turn" due to false positives. Labels are assigned based on (a) highest cosine similarity and (b) all labels with similarity $> 0.2$.

\subsection{Context}
\begin{figure}[htb]
    \centering
    \centerline{
    \includegraphics[width=0.45\columnwidth]{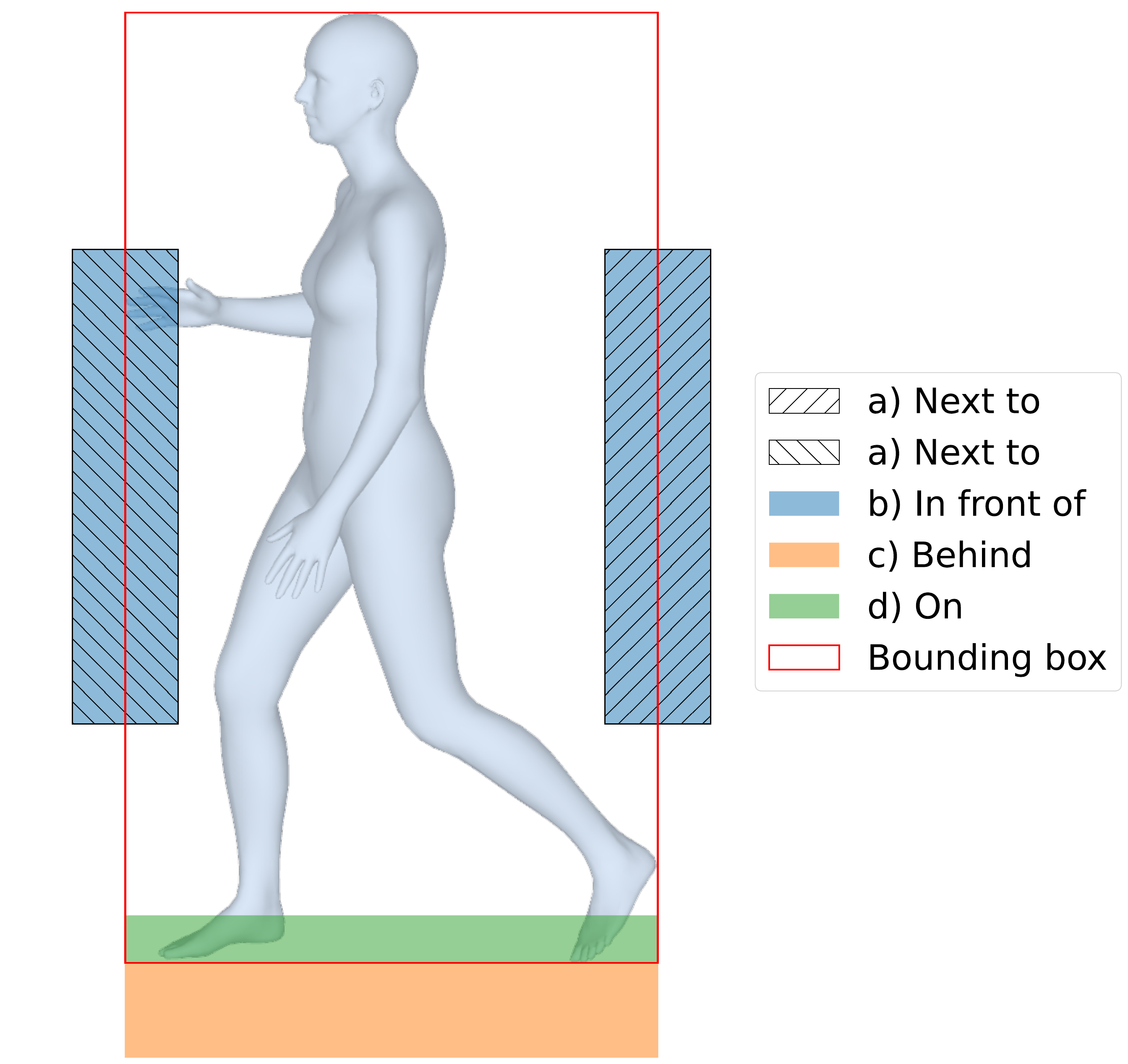}
    }
    \caption{Illustration of spatial‐relational context regions around a person’s bounding box (red): a) “on”, b) “behind”, c) “in front of”, and d) “next to”}
    \label{fig:spat-rel}
\end{figure}

\begin{figure}[htb]
  \centering
  \subfloat[\textbf{Next to} a bus\label{fig:sr-image1}]{
    \includegraphics[width=0.45\linewidth]{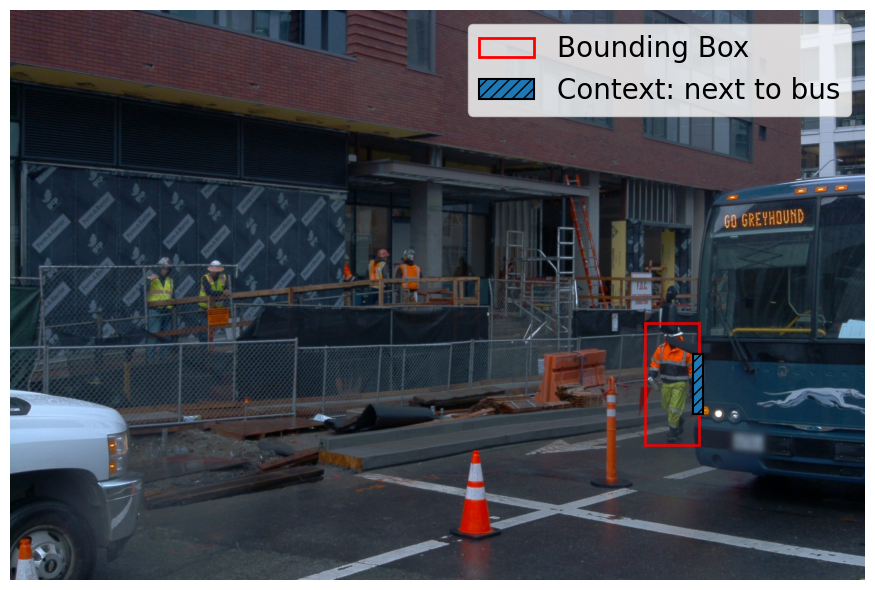}
  }
 \subfloat[\textbf{In front of} a car\label{fig:sr-image4}]{
    \includegraphics[width=0.45\linewidth]{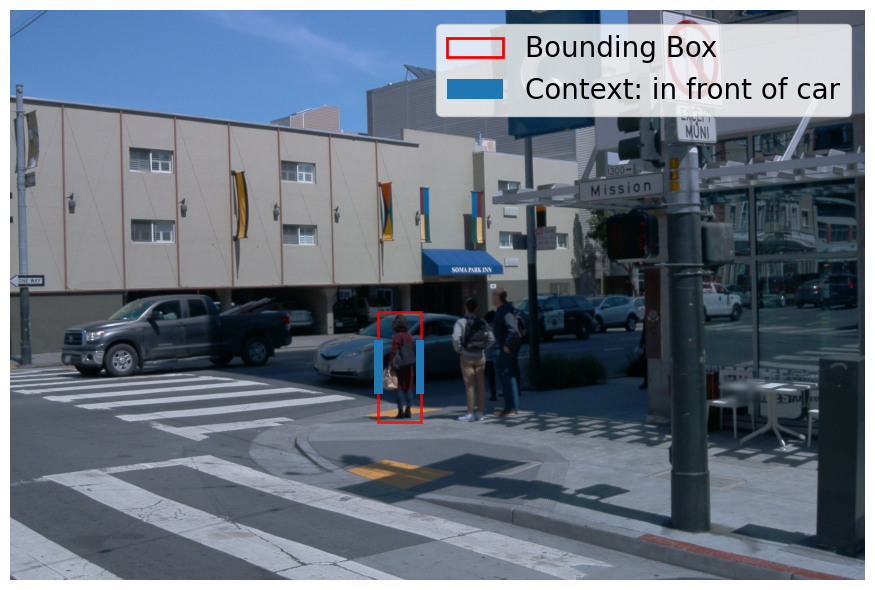}
  }\\[0ex]
  \subfloat[\textbf{Behind} a police car\label{fig:sr-image2}]{
    \includegraphics[width=0.45\linewidth]{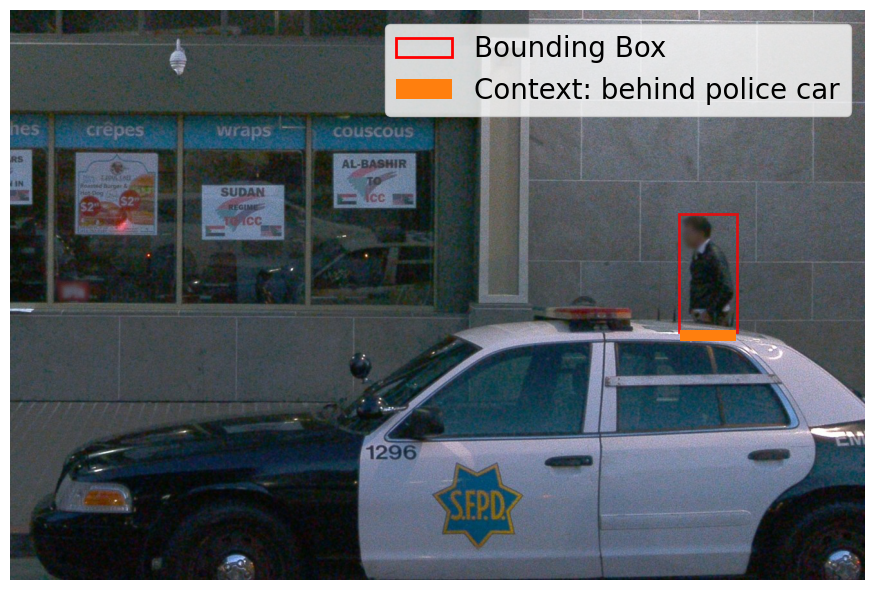}
  }
  \subfloat[\textbf{On} a sidewalk\label{fig:sr-image3}]{
    \includegraphics[width=0.45\linewidth]{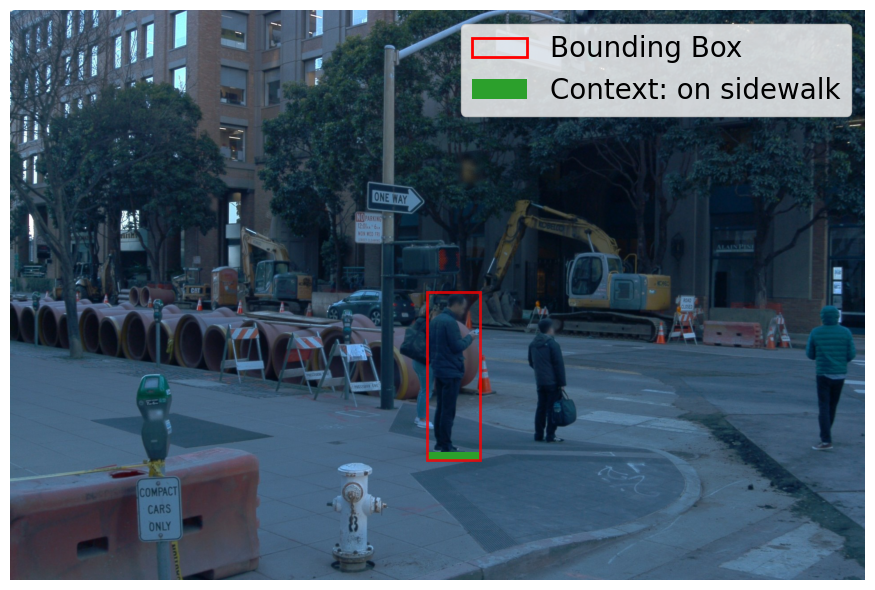}
  }

  \caption{Dataset examples of all four automatically labeled spatial relationships }
  \label{fig:spatial_relationships}
\end{figure}

In addition to labeling human motions, we annotate each person’s context based on their surroundings, using semantic segmentation and relative positions in the image. By leveraging the predominantly planar nature of traffic environments, we can define spatial relationships with respect to common scene elements.
For label generation, we consider two categories of contextual information: 
1)~Positional ground-level context — indicating the specific ground surface a person is located on (e.g., on the road, on a crosswalk, or on the sidewalk). 
2)~Spatial-relational context — capturing the relative positioning of people with respect to dynamic and static scene elements, such as vehicles, buildings, and other infrastructure.
We derive four spatial relationships, namely a) "next to", b) "in front of", c) "behind" and d) "on". Relationships are derived based on bounding box information from the Waymo Open Dataset (WOD), as illustrated in Figure~\ref{fig:spat-rel}. 
To obtain both the positional ground-level context and spatial-relational context classes, we use pretrained ODISE \cite{xu2023open} and ONEFORMER \cite{jain2023oneformer} models to produce two distinct semantic segmentations for each image.
ODISE is a pretrained open-vocabulary segmentation model that enables the incorporation of novel classes without the need for retraining. We primarily use COCO \cite{lin2014microsoft} and ADE20K \cite{zhou2019semantic} class labels, along with additional autonomous driving categories such as "e-scooter", "driveway", "police car", "fire truck", and "ambulance".
Since segmentation models such as ODISE struggle to distinguish between crosswalks and streets, we instead utilize the ONEFORMER model trained on the Mapillary Vistas dataset~\cite{neuhold2017mapillary} to segment our images for accurate ground-level context.
Both models segment the middle frame of each sequence. 
\\
For spatial-relational context, we use non-ground object segmentation from the ODISE model to identify nearby scene elements and determine persons' relative positions as follows (cf. Figure \ref{fig:spat-rel}):
a)~A person is labeled as "next to" an object if the object appears on either the left or right side of their bounding box. Specifically, we consider the area between 5\% inside and 5\% outside the box horizontally, and between 25\% and 75\% of its height vertically (shown as the striped blue regions in Figure \ref{fig:spat-rel}). An example scene depicting a person next to a bus is given in Figure \ref{fig:sr-image1}. 
b)~A person is labeled as "in front of" an object if the same object class appears on both sides of the bounding box (i.e. in the striped blue areas of Figure \ref{fig:spat-rel}). Figure \ref{fig:sr-image4} presents a scene where a person is standing in front of a car.
c)~A person is labeled as "behind" an object if relevant classes within a region extending 10\% are below the bounding box (orange area in Fig.~\ref{fig:spat-rel}). Figure~\ref{fig:sr-image2} shows an example of a person behind a police car.

To identify the positional ground-level context, we use ONEFORMER to extract semantic class labels from the pixels within the bottom 5\% of the person’s bounding box (green area in Figure~\ref{fig:spat-rel}).
This enables us to determine whether the person is on the road, a crosswalk, or another surface. If no clear ground class is detected, we fall back to segmentation with ODISE. For this context, we assign only a single class label. An example is given in Figure \ref{fig:sr-image3}, where a person on a sidewalk is correctly labeled and localized on a sidewalk.

\subsection{Combining Motion and Context} 
Once the motion and context for each person are defined, we combine them into two types of annotations. The simple form concatenates a motion and context word, resulting in labels like \{motion\} \{context\} (e.g., "walking crosswalk"). Since multiple motions and contexts may apply, this yields up to $c \times m$ combinations per person.\\
Additionally, we create more detailed sentence-style annotations of the form \{person\} \{motion\} \{relation\} \{context\}, using four synonyms for \{person\} ("A person", "Someone", "Somebody" and "A pedestrian") and all valid motions and contexts (e.g., "A person is waving on a crosswalk", "Someone is walking behind a car"). Thus, annotations increase combinatorially with multiple motions, contexts, and synonyms. We also incorporate approximately 30 synonym terms for motion and context labels (e.g., “street” and “road”), effectively doubling the number of annotations for each alternative.

\section{Method}
\label{method:model}
\begin{figure}[htb]

    \centerline{
    \includegraphics[width=1.0\columnwidth]{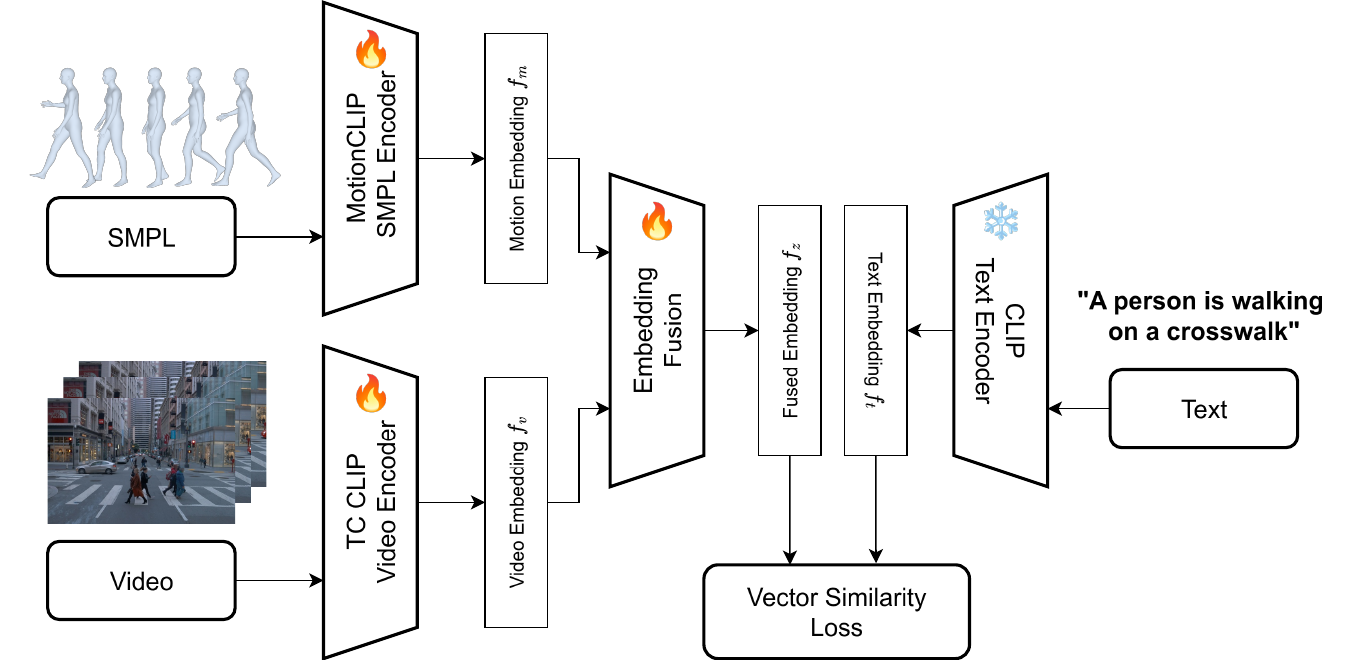}
    }
    \caption{ContextMotionCLIP architecture: SMPL sequences and video are encoded via the MotionCLIP SMPL Encoder and TC CLIP Video Encoder, fused by the Embedding Fusion module, and aligned with the output of the CLIP Text Encoder using a Vector Similarity Loss.}
    \label{fig:arch}
\end{figure}

Our model retrieves human motions over SMPL estimates along with their contextual information over video data. The overall architecture is illustrated in Figure~\ref{fig:arch}. We utilize MotionCLIP and TC-CLIP as motion and video encoders, respectively. Their fused outputs form a unified representation, trained to align with the text embedding of the corresponding annotation.

\subsection{Model Selection}
Existing models cannot jointly process motion and visual context. To enable retrieval based on both in large datasets, we combine encoded SMPL sequences with encoded visual context from video.
For the motion backbone, we use MotionCLIP, originally trained on AMASS with BABEL text annotations. We retrain it at 10~Hz to match the WOD frame rate and modify the loss to emphasize text–motion similarity over image–motion similarity.
Unlike MotionCLIP, which uses a 6D continuous rotation format, our SMPL data loader converts motion from the neutral body model into joints represented in XYZ coordinates.

As our visual backbone, we use TC-CLIP, which achieves state-of-the-art performance on several video recognition benchmarks \cite{kim2024tcclip} and uses a CLIP-based text encoder, making it well-suited for our natural language descriptions.
We disable the video-conditional prompting module to keep the video and text encoders fully decoupled, and apply the video augmentation strategies proposed by TC-CLIP during training. 
Additionally, we draw a red bounding box with a 2-pixel stroke-width around each person in every input frame, producing one video per person. This leverages observations from~\cite{shtedritski2023does} that CLIP-like models are sensitive to simple graphical cues such as colored shapes.

\subsection{Feature Fusion}
The final layer of our model fuses embeddings from MotionCLIP (\( \mathbf{f}_m \)) and TC-CLIP (\( \mathbf{f}_v \)), both in \( \mathbb{R}^{512} \), into a unified representation \( \mathbf{f}_z \in \mathbb{R}^{512} \), aligned with the dimensionality of the CLIP text encoder output \( \mathbf{f}_t \) \( \in \mathbb{R}^{512} \) from \cite{radford2021learning}. 
We explore three combination strategies: simple concatenation (Eq.~\ref{eq:concat}), bilinear pooling (Eq.~\ref{eq:bilinear}), and self-attention over the two concatenated modality features (Eq.~\ref{eq:attn}), which are jointly processed using multi-head attention. The output is mean-pooled to obtain a fixed-size fusion vector.
\begin{alignat}{2}
\mathbf{f}_{\text{concat}}  &= [\mathbf{f}_m;\, \mathbf{f}_v]\label{eq:concat}  \\
\mathbf{f}_{\text{bilinear}} &= \mathbf{f}_m \otimes \mathbf{f}_v \label{eq:bilinear} \\
\mathbf{f}_{\text{attn}}    &= \text{MeanPool}(\text{SelfAttn}([\mathbf{f}_m;\, \mathbf{f}_v]))  \label{eq:attn}
\end{alignat}
In all cases, the resulting representation passes through layer normalization (\( \mathcal{L} \)) and dropout (\( p=0.5 \), \( \mathcal{D} \)), followed by a 512-unit MLP (\( \mathcal{M} \)) to produce the final embedding \( \text{f}_{z} \):
\begin{align}
\mathbf{f}_{\text{z}} &= \mathcal{M}\bigl(\mathcal{D}(\mathcal{L}(\mathbf{f}_k))\bigr), \\[6pt]
\quad \mathbf{f}_k &\in 
\left\{\,
  \mathbf{f}_{\text{concat}}\in\mathbb{R}^{1024},\;
  \mathbf{f}_{\text{bilinear}}\in\mathbb{R}^{512^2},\;
  \mathbf{f}_{\text{attn}}\in\mathbb{R}^{512}
\right\}
\end{align}

\subsection{Training}
MotionCLIP and TC-CLIP backbones are trained jointly with our fusion module on our WayMoCo dataset, while the pre-trained CLIP text encoder remains frozen throughout training.
With a batch size of six, we optimize with AdamW~\cite{loshchilov2018decoupled} (initial learning rate $10^{-5}$, decayed exponentially to $10^{-6}$ over 50 epochs) and employ gradient scaling via GradScaler~\cite{micikevicius2018mixed} to stabilize training and prevent exploding gradients.
All feature vectors are \( \ell_2 \)-normalized prior to loss computation.

The fused representation $\text{f}_{z}$ is trained to match the output of a CLIP text encoder $\text{f}_{t}$ (\texttt{ViT-B/32} configuration). To achieve this, we experiment with several loss functions inspired by state-of-the-art methods: Cosine similarity loss \( \mathrm{L}_{\text{cos}} \) (Eq.~\ref{eq:lcos}) as deployed in MotionCLIP~\cite{tevet2022motionclip}, InfoNCE \( \mathrm{L}_{\text{InfoNCE}} \) (Eq.~\ref{eq:linfonce}) as adopted by TMR~\cite{petrovich2023tmr}, and soft target cross-entropy \( \mathrm{L}_{\text{soft}} \) (Eq.~\ref{eq:lsoft}) as used by TC-CLIP~\cite{kim2024tcclip}.
\begin{align}
\mathrm{L}_{\text{cos}} &= 1 - \frac{\mathbf{f}_z \cdot \mathbf{f}_t}{\lVert \mathbf{f}_z \rVert \; \lVert \mathbf{f}_t \rVert} \label{eq:lcos} \\
\mathrm{L}_{\text{soft}} &= - \sum_{i} [\mathbf{f}_t]_i \log [\mathbf{f}_z]_i \label{eq:lsoft} \\
\ell_i &= \log \left( \frac{\exp\left( (\mathbf{f}_z^i)^\top \mathbf{f}_t^i / \tau \right)}{\sum_j \exp\left( (\mathbf{f}_z^i)^\top \mathbf{f}_t^j / \tau \right)} \right) \nonumber \\
\mathcal{L}_{\text{InfoNCE}} &= -\frac{1}{2N} \sum_i \ell_i \quad \text{with } \tau = 0.5
\label{eq:linfonce}
\end{align}

\subsection{Querying Framework}
\begin{figure}[htb]
    \centerline{
    \includegraphics[width=0.95\columnwidth]{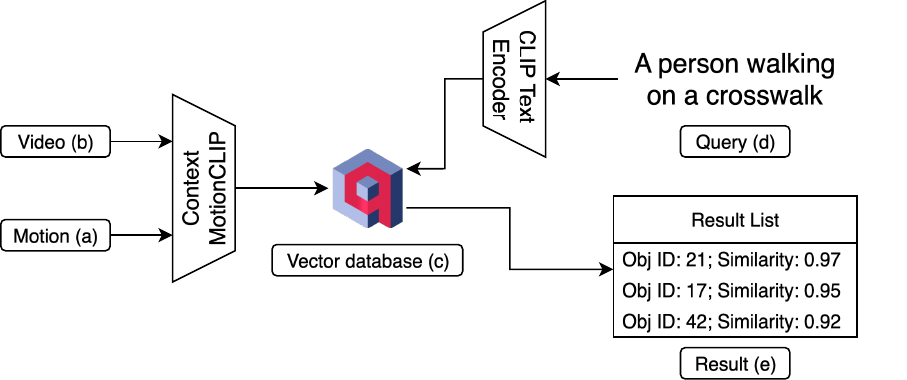}
    }
\caption{Overview of our context–motion retrieval framework. Motion (a) and video (b) are encoded by ContextMotionCLIP, indexed in a vector database (c) \cite{qdrant2025}, and at query time a text description (d) is embedded and matched to produce a ranked list of object IDs (e) with cosine‐similarity scores.}
    \label{fig:database}
\end{figure}

We implement a framework to efficiently query data via text from a vector database, which stores all embeddings generated by our ContextMotionCLIP model for the test sets. The architecture is depicted in Figure \ref{fig:database}.
A server running the text encoder handles incoming queries and returns the top-$n$ matching scenes—including both the person and their context—within milliseconds. Because vector databases scale to billions of embeddings, this approach imposes virtually no limit on future dataset size, enabling fast, open-vocabulary retrieval of context-based motions. It also allows direct qualitative exploration of results by visually inspecting retrieved scenarios for arbitrary natural language queries.

\section{Experiments and Results}

\subsection{Dataset}
Our WayMoCo dataset comprises 27,466 \textit{valid} sequences. Table \ref{tab:dataset-count} reports the numbers of samples in the training (68.5\%), validation (13\%) and test (18.4\%) splits, as well as the annotation counts under two settings: setting 1 contains the original labels, while setting 2 (*) additionally includes synonyms augmenting the ground-truth annotations. Because multiple motion and context labels can exist for each video sequence, the dataset includes combinations of all possible pairs for each sequence. All combinations are included in the synonym-enriched full-sentence annotations.

\begin{table}[htb]
\caption{Overview of dataset splits, number of sequences and counts of motion, context, combined, and sentence annotations, with (*) and without synonym augmentation.}
  \centering
  \begin{tabular}{l c c c c c}
    \toprule
    \textbf{Split} & \textbf{Samples} & \textbf{Motion} & \textbf{Context} & \textbf{Combination} & \textbf{Sentences} \\
    \midrule
    Train             & 18,814 & 21,966 & 39,185 & 46,067 & 46,067 \\
    Val               & 3,585  & 4,212  & 7,413  & 8,765  & 8,765  \\
    Test              & 5,067  & 5,955  & 10,149 & 12,066 & 12,066 \\
    \midrule
    \textbf{Total}    & \textbf{27,466} & \textbf{32,133} & \textbf{56,747} & \textbf{66,898} & \textbf{66,898} \\
    \midrule
    Train*            & 18,814 & 33,189 & 42,115 & 74,514 & 412,380 \\
    Val*              & 3,585  & 6,273  & 7,937  & 13,970 & 77,632 \\
    Test*             & 5,067  & 9,092  & 11,056 & 19,929 & 110,240 \\
    \midrule
    \textbf{Total*}   & \textbf{27,466} & \textbf{48,554} & \textbf{61,108} & \textbf{108,413} & \textbf{600,252} \\
    \bottomrule
  \end{tabular}
  \label{tab:dataset-count}
\end{table}

\begin{figure}[htb]
    \centerline{
    \includegraphics[width=0.99\columnwidth]{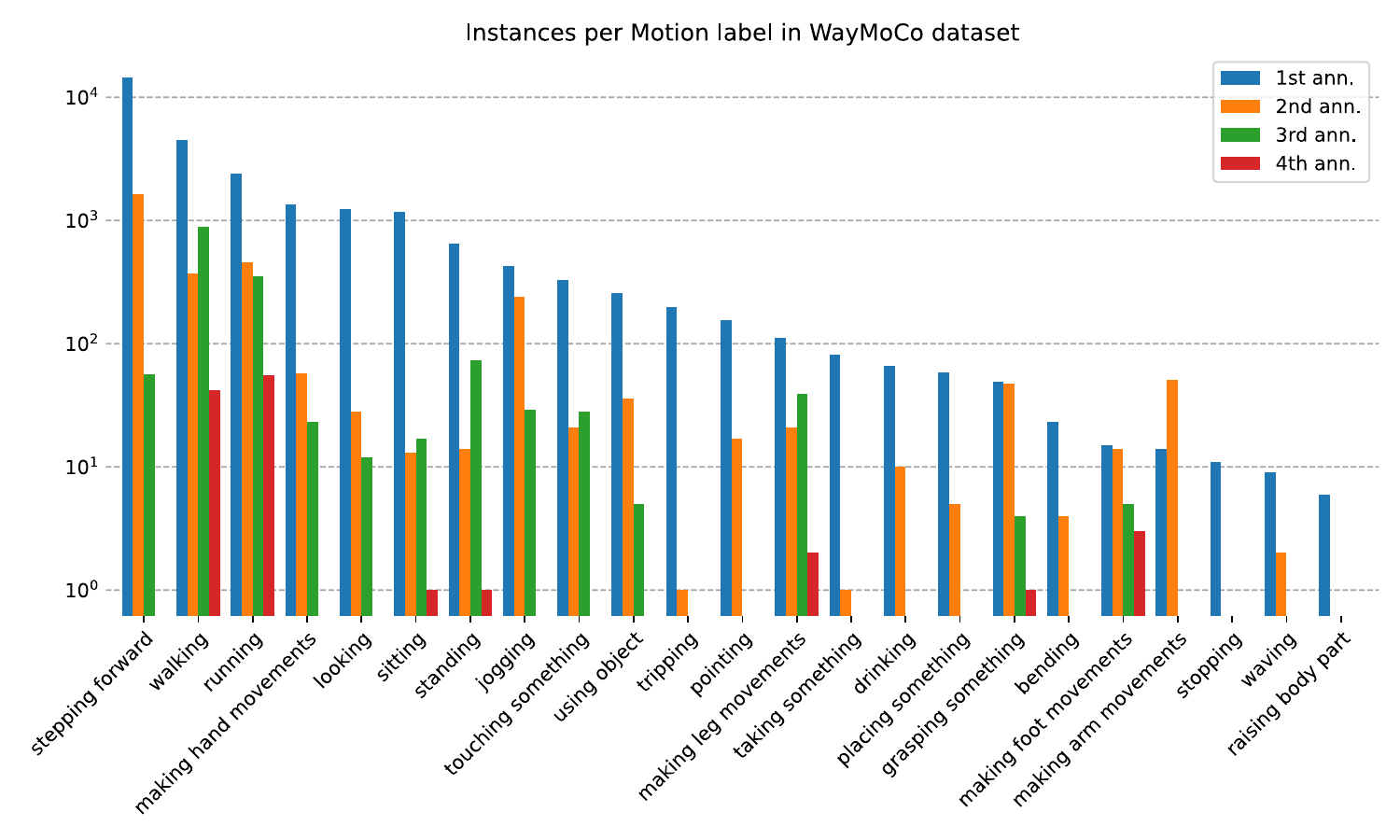}
    }

    \caption{Distribution of motion annotations. The chart shows how often each motion word was assigned, also indicating whether it was the 1st to 4th best-matching label for each sequence.}
    \label{fig:dist-motion}
\end{figure}

Figure~\ref{fig:dist-motion} shows the distribution of motion annotations. The chart groups motion labels by their rank as the top-1 to top-4 annotation for each sequence. The most common actions are stepping forward, walking, and running, which primarily describe overall body movement. In contrast, explicit hand motions like waving are rare and often subsumed under more generic labels such as making hand movements, rather than being distinctly annotated.

\begin{figure}[htb]
    \centering
        \centerline{
        \includegraphics[width=1.0\columnwidth]{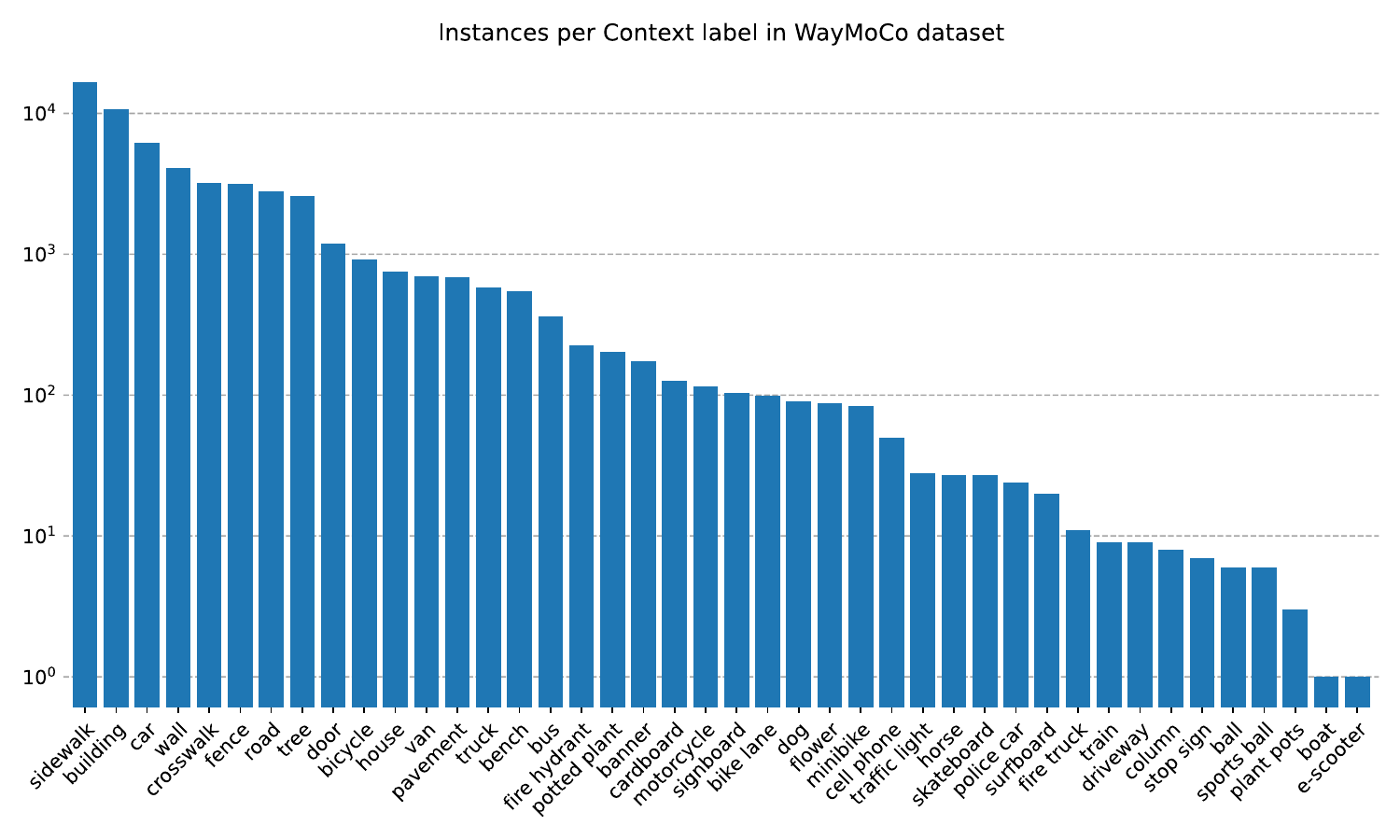}
        }
        \caption{Distribution of context annotations, showing how frequently each context label appears in our dataset.}
        \label{fig:dist-context}
\end{figure}

Figure~\ref{fig:dist-context} presents the distribution of context labels, highlighting the prevalence of sidewalk and crosswalk among ground-level categories. High counts are also seen for spatial relations involving nearby objects such as cars and buildings. Notably, ODISE’s open-vocabulary segmentation enables the correct annotation of rare classes like police car and fire truck, which are not part of its original training set.

Additionally, we evaluate the number of samples in each of the four spatial-relation categories: "on" (23,451), "behind" (16,592), "in front of" (11,546), and "next to" (5,158). Though not uniform, the distribution shows all spatial relations are sufficiently represented.

\subsection{Metrics}
To evaluate the performance of our motion scene retrieval method, we use top-k ranking, a standard measure in scene retrieval tasks \cite{humanml3d, kim2024tcclip, hanoonavificlip}. This metric indicates the percentage of cases where a correct retrieval appears among the top k ranked candidates.

\subsection{Implementation Details}

We partition the WOD dataset by splitting its original training set into new training and validation subsets. Specifically, sequences beginning with frames 1–6 are assigned to the training subset, and those beginning with frames 7–9 to the validation subset. We then treat the WOD validation set as our test set. Since WOD does not provide annotations for its official test set, that portion cannot be utilized. All subsequent experiment results are reported on our test dataset. Unless stated otherwise, we use the first annotation according to our dataset for each sample in the form of \{motion\} \{context\}.

To adapt to the WOD’s 10~Hz frame rate (versus MotionCLIP’s original 30~Hz) and 20‑frame sequence lengths, we retrain MotionCLIP’s SMPL encoder with TC‑CLIP’s video encoder. Due to the lack of reproducibility that hinders the reconstruction of intermediate frames in MotionCLIP, we opt to use repeated frames. We set the loss weights to \(\lambda_{\text{image}}=0.1\) and \(\lambda_{\text{text}}=1.0\).
We initialize the motion backbone with our newly trained MotionCLIP and the visual backbone with the pretrained zero-shot configuration of TC-CLIP. Results are reported after 25 epochs of training, as we observe overfitting beyond that point.

Experiments are performed on a single NVIDIA RTX 4090 GPU, except for bilinear and self-attention fusion, which requires higher memory and was trained on an A6000.

\subsection{Qualitative results}
\begin{figure}[htb]
  \centering
  \makebox[\columnwidth][c]{%
    \includegraphics[width=0.328\columnwidth]{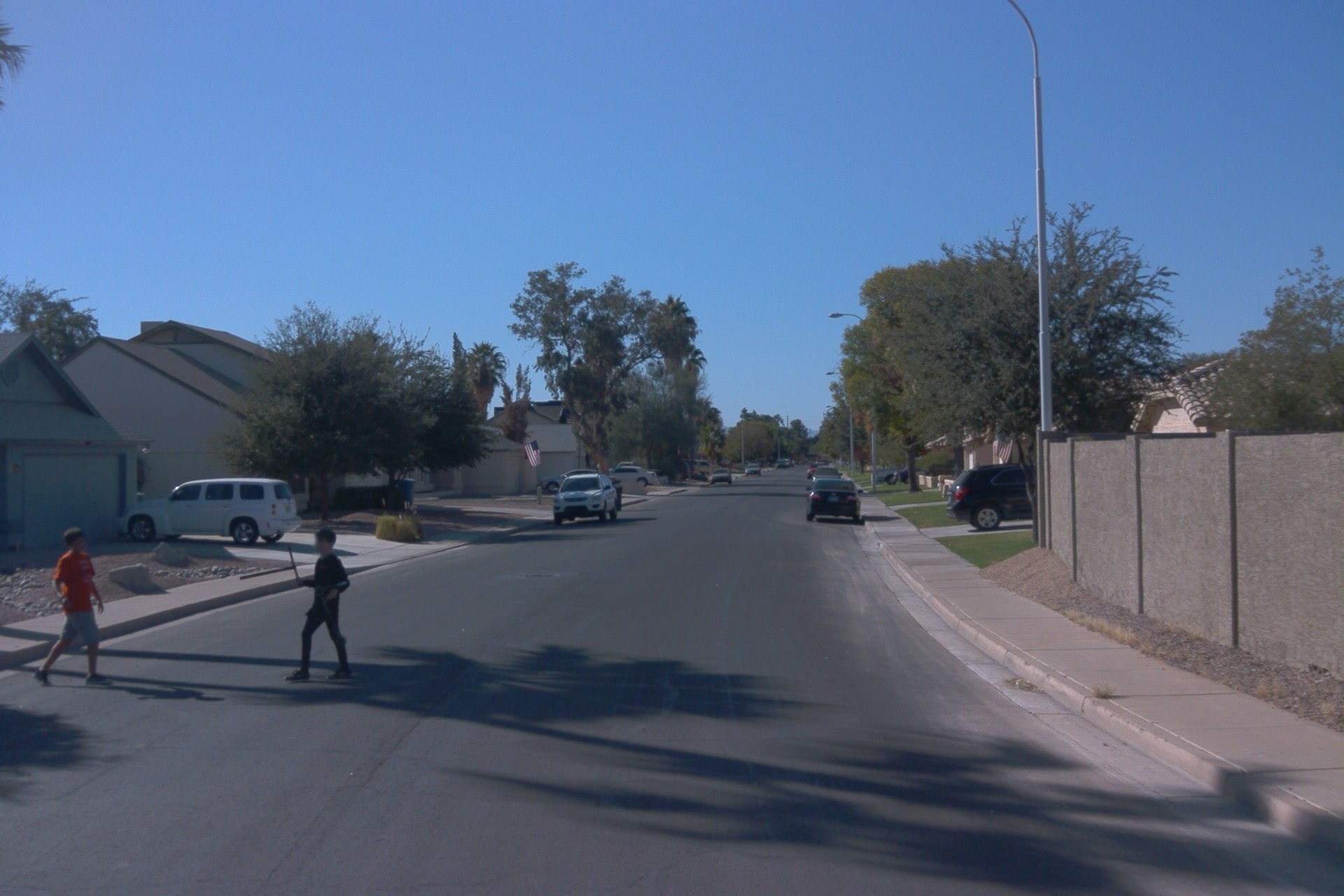}
    \includegraphics[width=0.11\columnwidth]{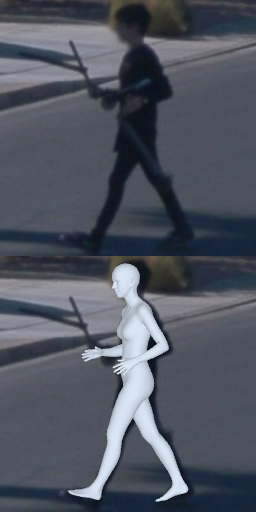}
    \includegraphics[width=0.11\columnwidth]{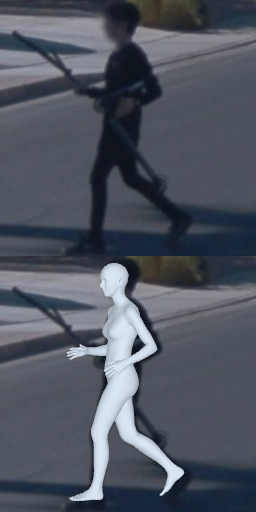}
    \includegraphics[width=0.11\columnwidth]{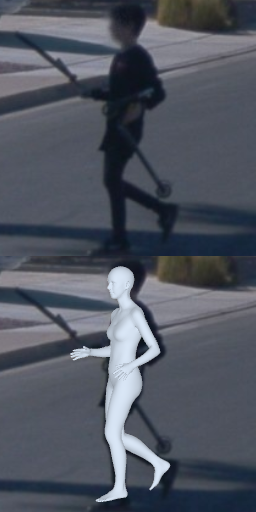}
    \includegraphics[width=0.11\columnwidth]{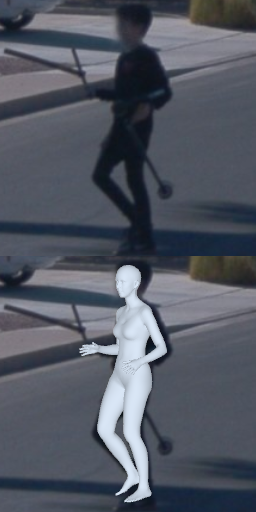}
    \includegraphics[width=0.11\columnwidth]{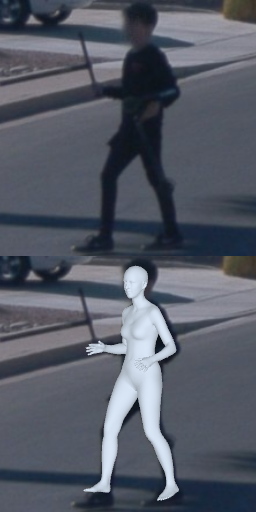}
  }
  \caption{Qualitative example of motion annotation. Left: full scene frame. Right: consecutive SMPL overlays showing a “walking” sequence.}
  \label{fig:motion-walking}
\end{figure}

To evaluate the quality of our automatic context annotations, we manually inspect various examples from the dataset. In most cases, the assigned context classes and spatial-relation labels correctly match the visual scene content. 
We also qualitatively evaluate our SMPL extraction pipeline, both at the frame level and on assembled motion sequences. Figure \ref{fig:motion-walking} illustrates how TokenHMR reliably recovers accurate SMPL poses from individual video frames. When concatenated into two-second sequences, our MotionCLIP-based annotator correctly recognizes the actual human motion, assigning the annotation "walking", which aligns with human assessment.

\begin{figure*}[tb]
  \centering

    \includegraphics[width=0.49\columnwidth]{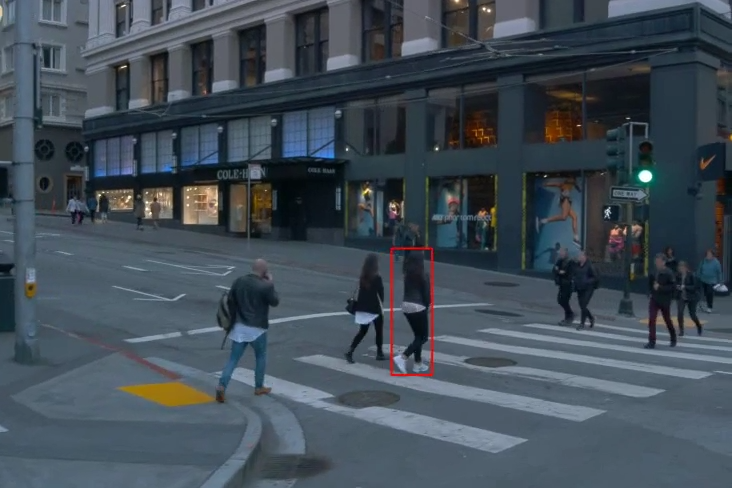}
    \hfill
    \includegraphics[width=0.49\columnwidth]{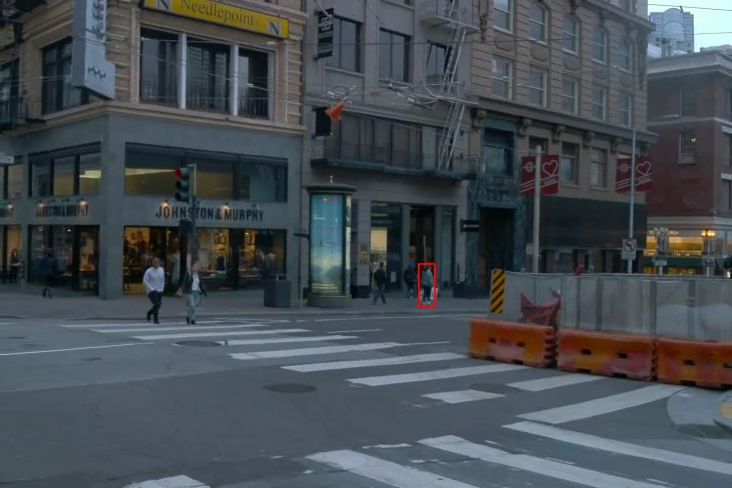}
    \hfill
    \includegraphics[width=0.49\columnwidth]{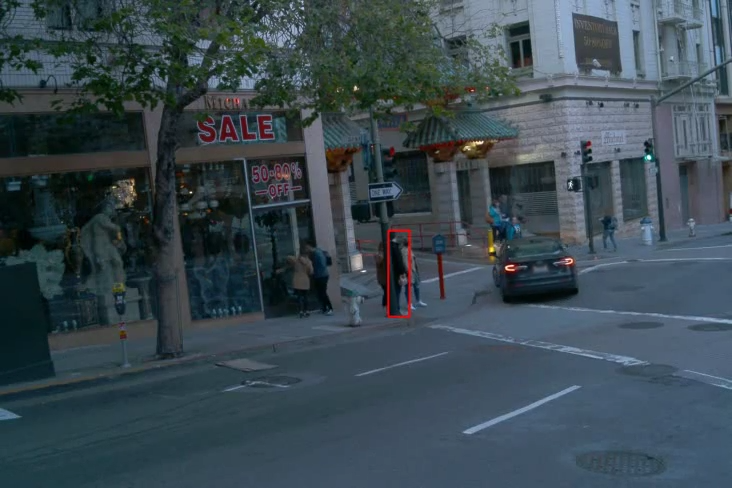}
    \hfill
    \includegraphics[width=0.49\columnwidth]{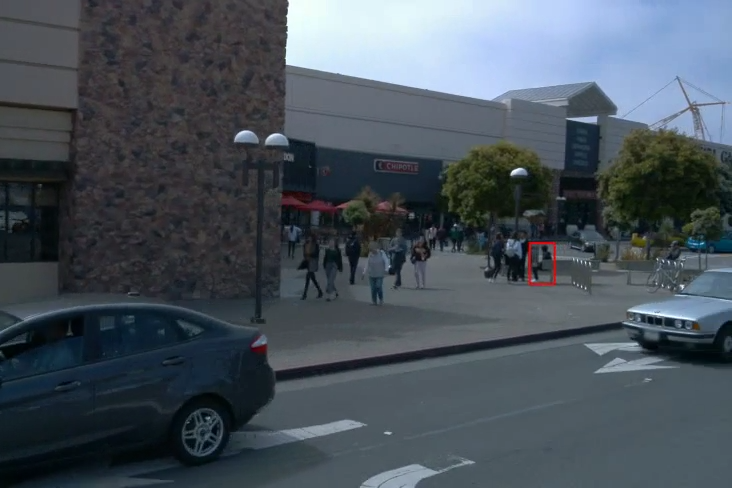}
    \\[0.5ex]

    \includegraphics[width=0.49\columnwidth]{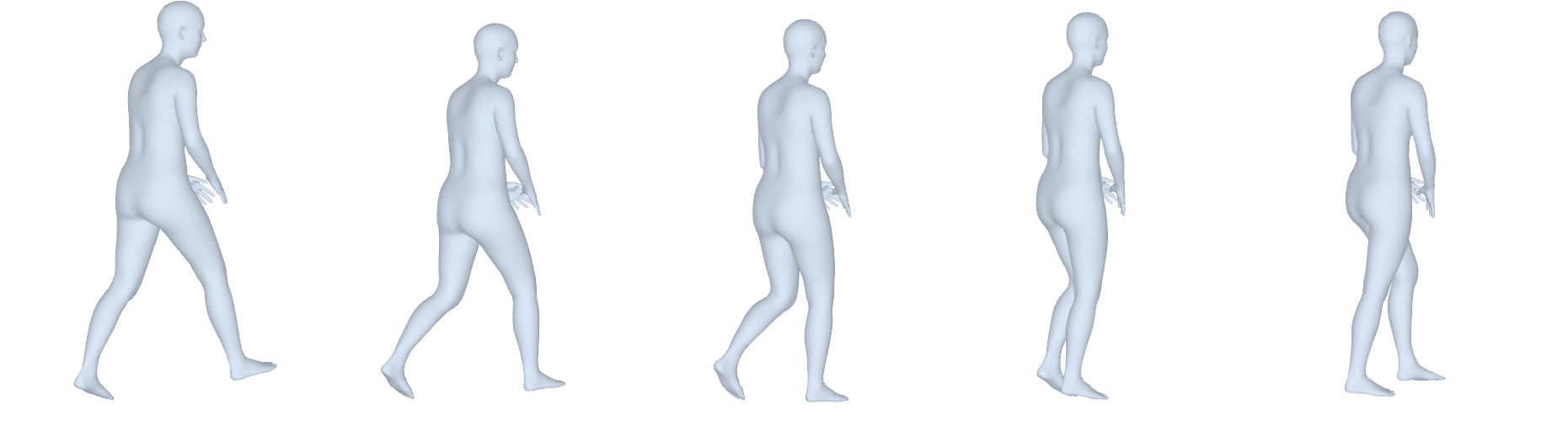}
    \hfill
    \includegraphics[width=0.49\columnwidth]{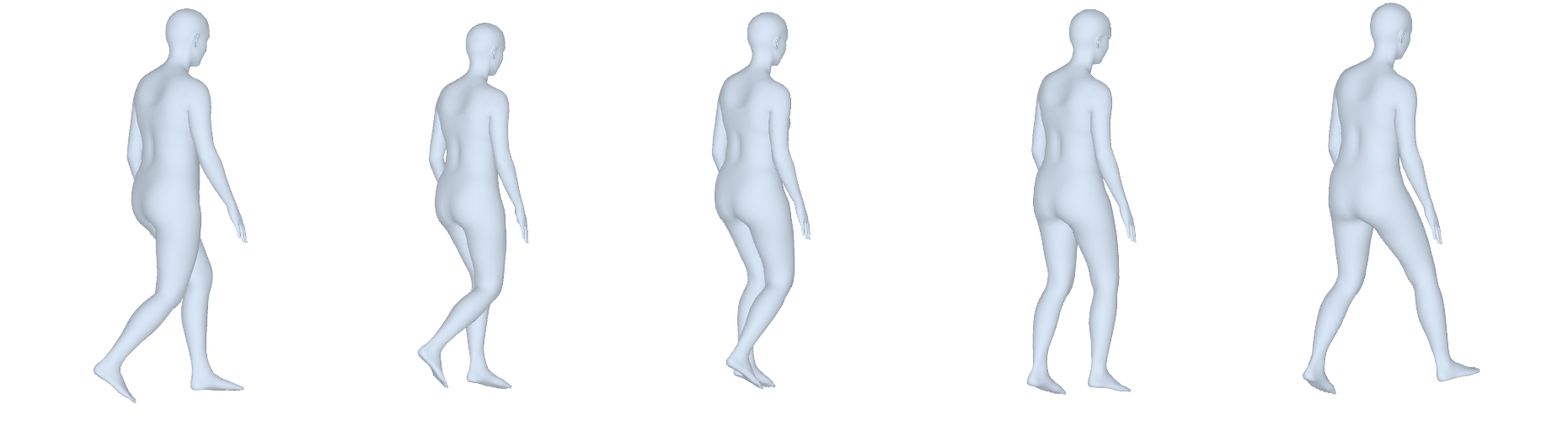}
    \hfill
    \includegraphics[width=0.49\columnwidth]{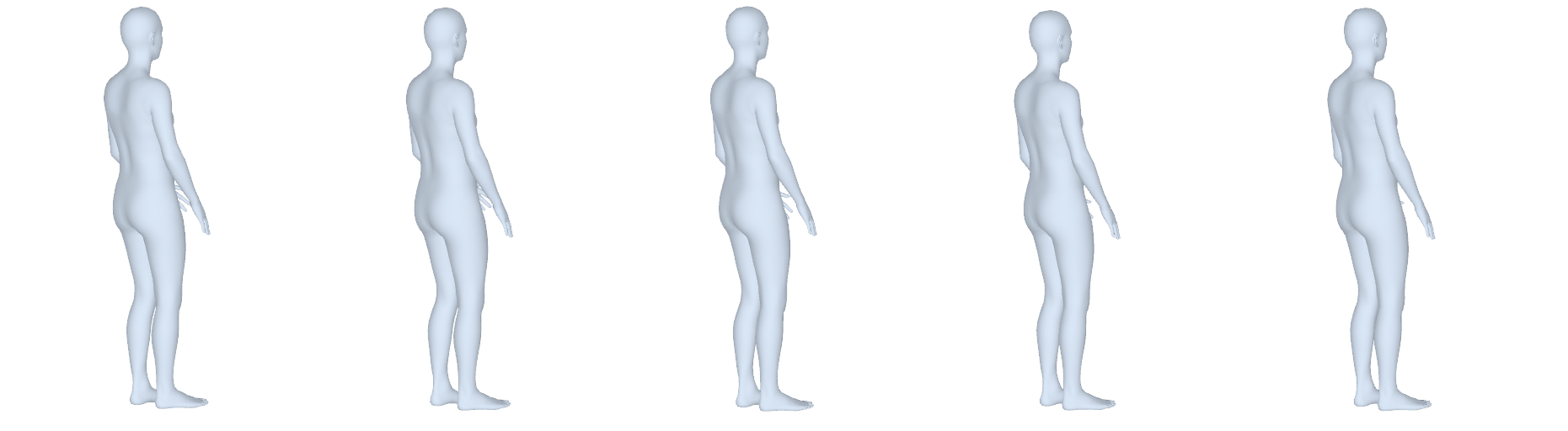}
    \hfill
    \includegraphics[width=0.49\columnwidth]{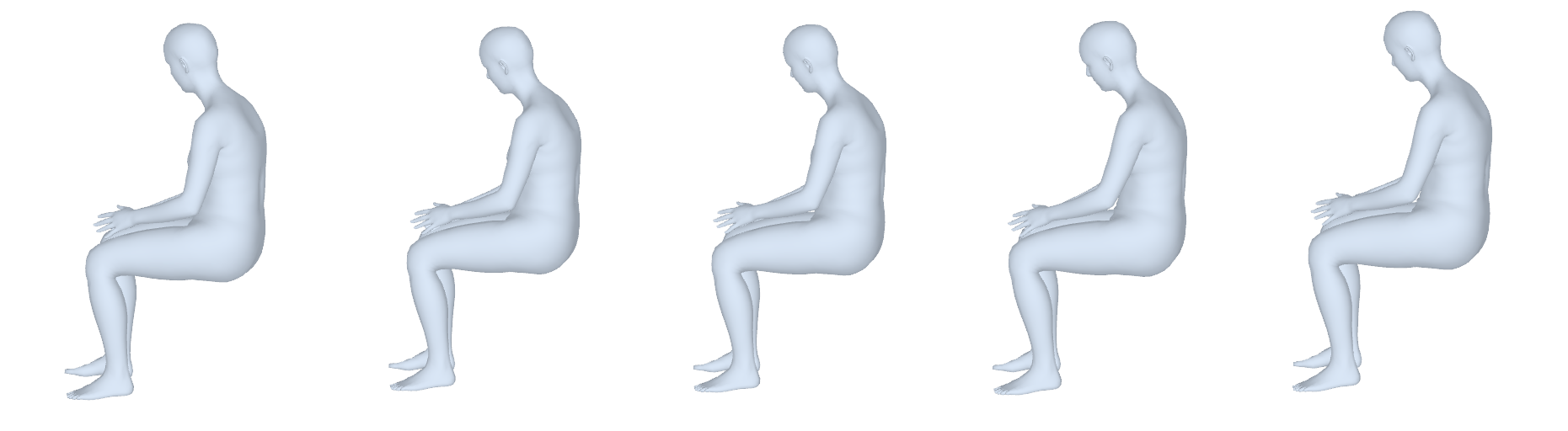}
    \\[1.5ex]
      
    \includegraphics[width=0.49\columnwidth]{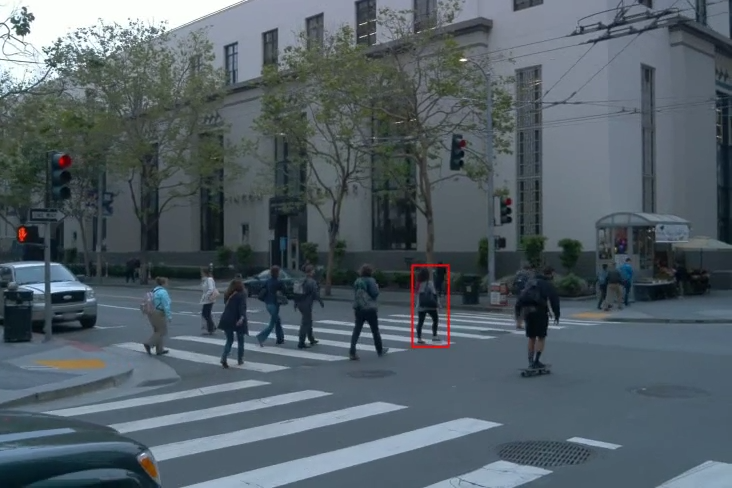}
    \hfill
    \includegraphics[width=0.49\columnwidth]{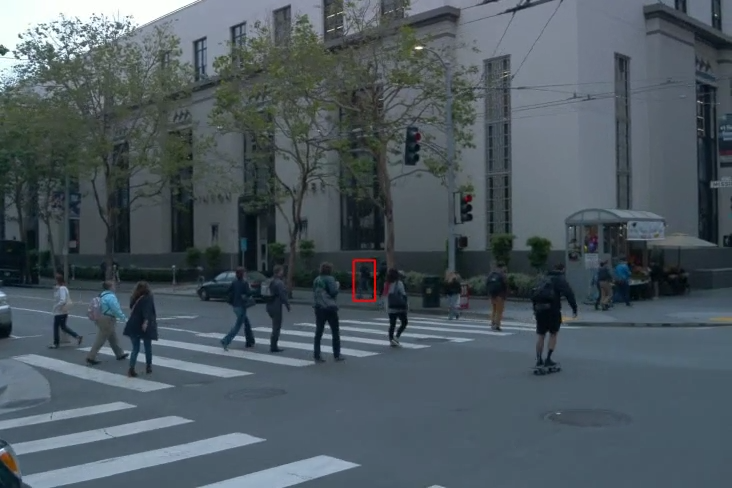}
    \hfill
    \includegraphics[width=0.49\columnwidth]{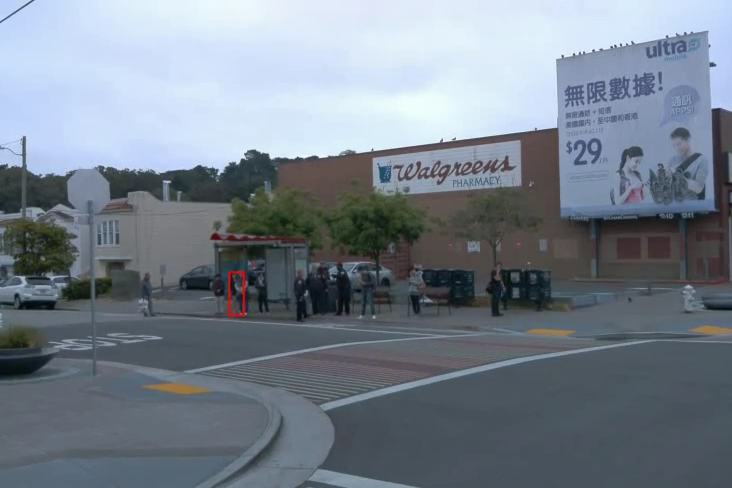}
    \hfill
    \includegraphics[width=0.49\columnwidth]{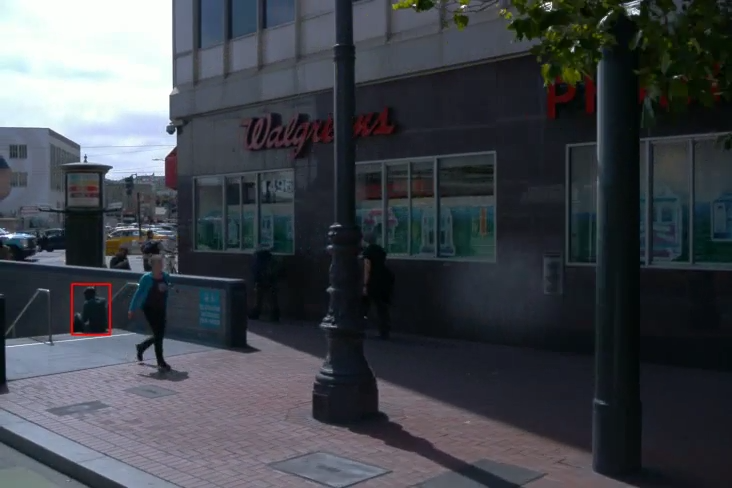}
    \\[0.5ex]
        
    \includegraphics[width=0.49\columnwidth]{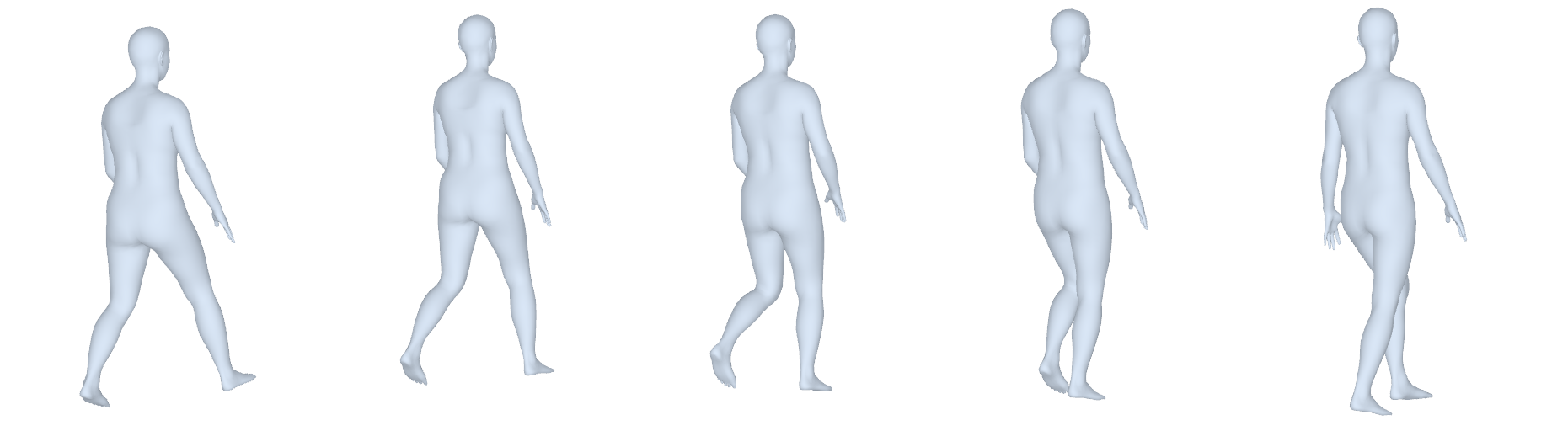}
    \hfill
    \includegraphics[width=0.49\columnwidth]{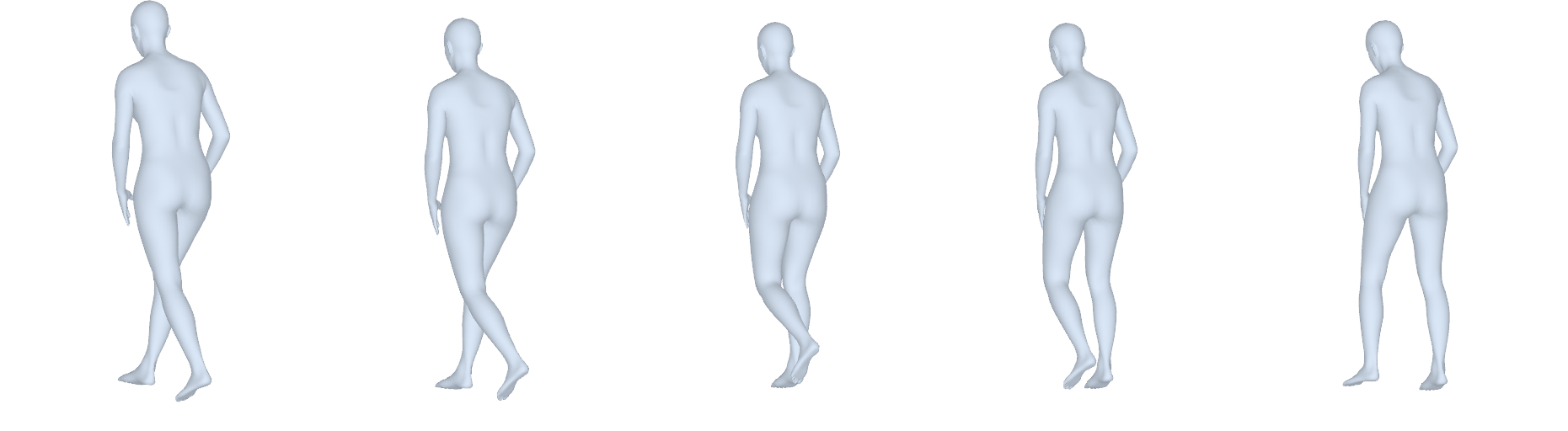}
    \hfill
    \includegraphics[width=0.49\columnwidth]{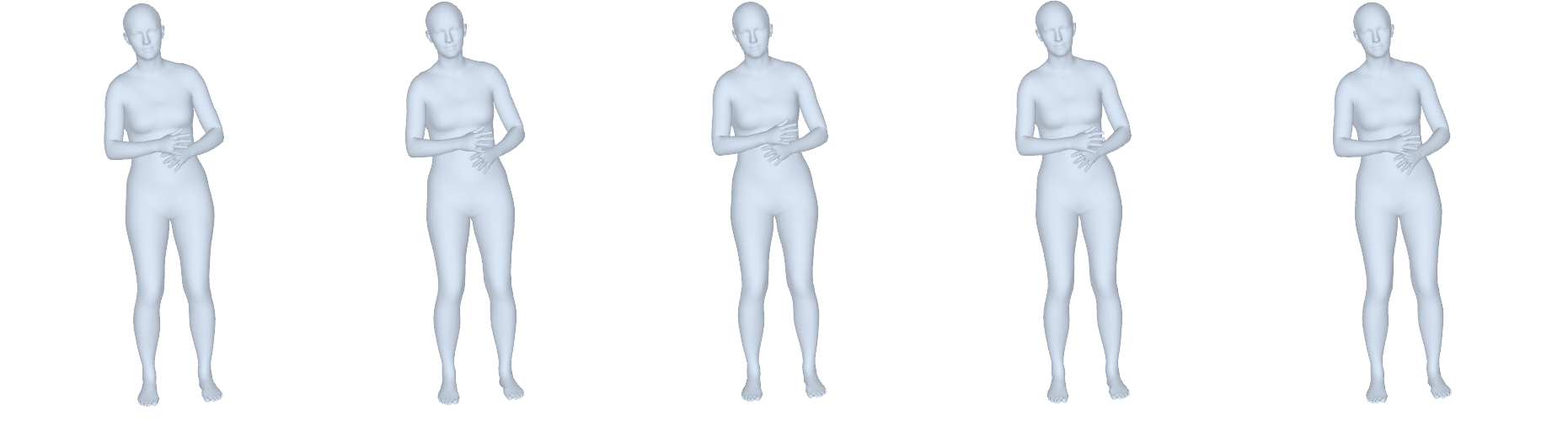}
    \hfill
    \includegraphics[width=0.49\columnwidth]{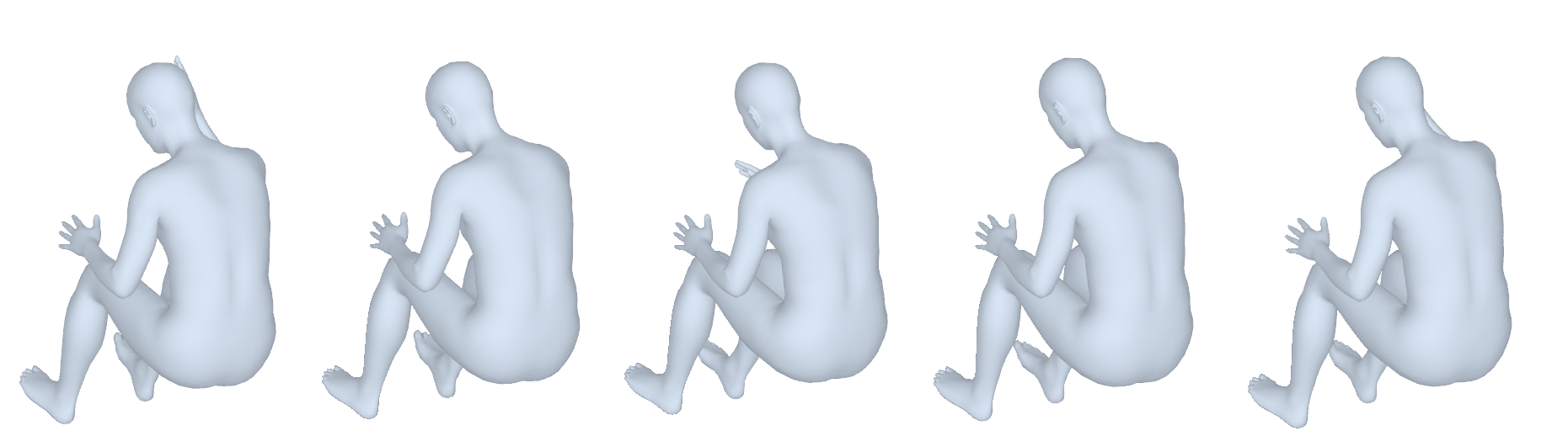}
    \\[1.5ex]

    \includegraphics[width=0.49\columnwidth]{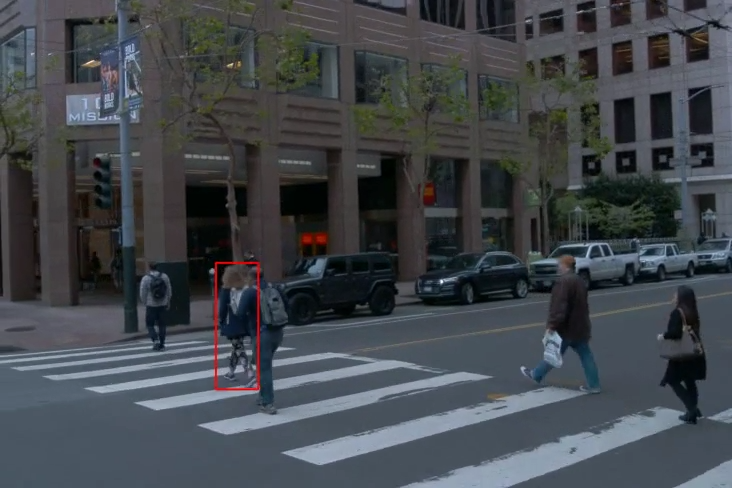}
    \hfill
    \includegraphics[width=0.49\columnwidth]{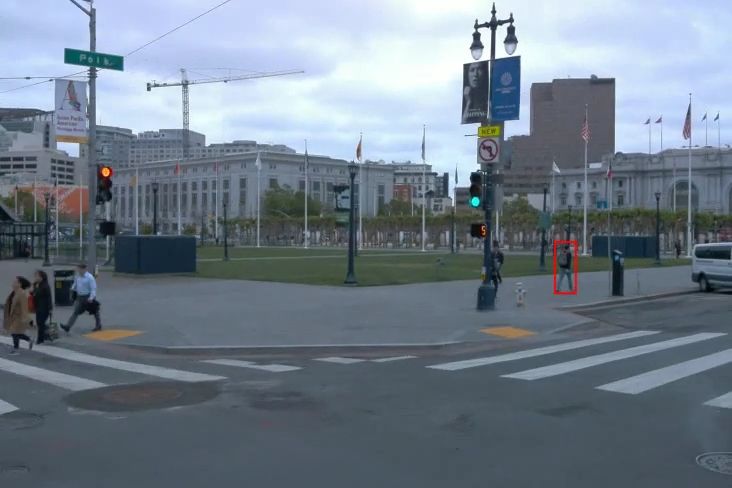}
      \hfill
    \includegraphics[width=0.49\columnwidth]{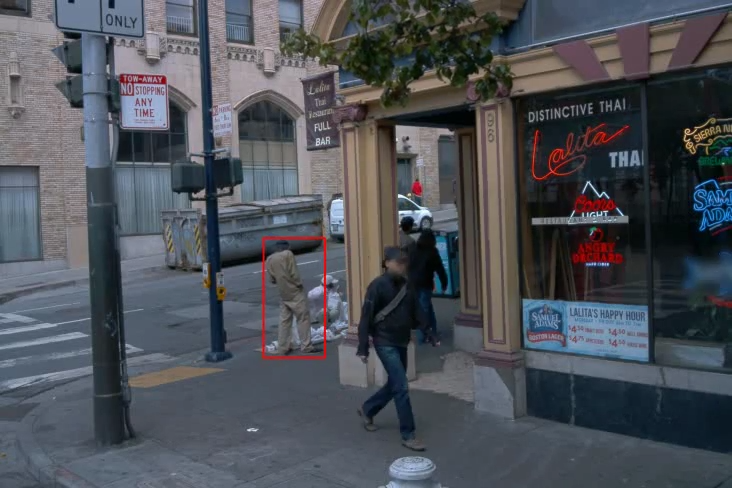}
    \hfill
    \includegraphics[width=0.49\columnwidth]{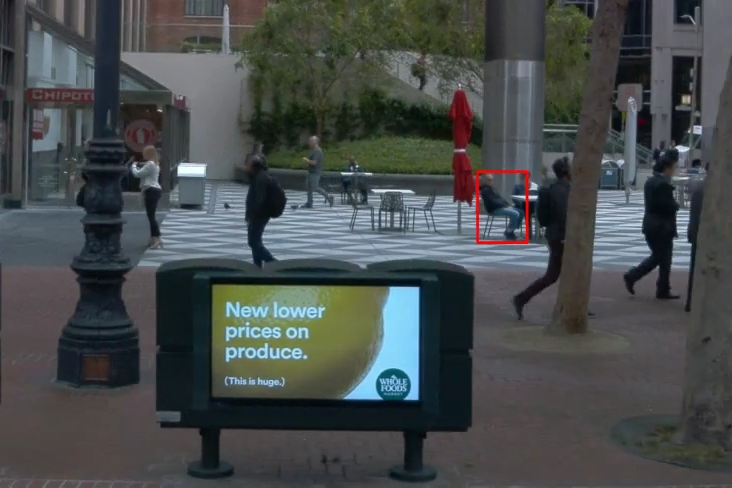}
    \\[0.5ex]

    \includegraphics[width=0.49\columnwidth]{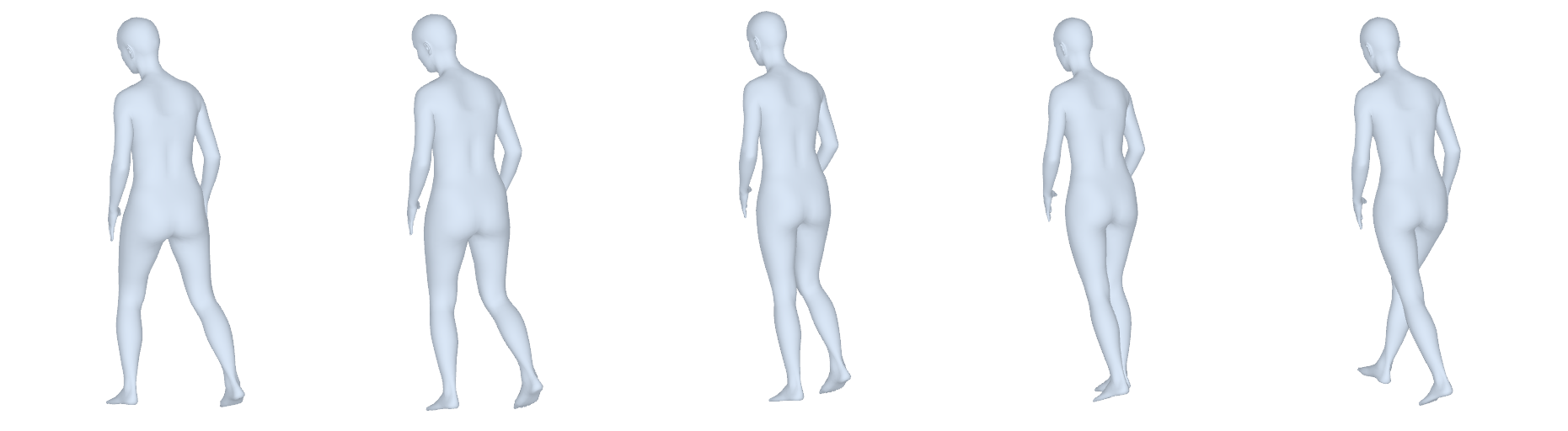}
    \hfill
    \includegraphics[width=0.49\columnwidth]{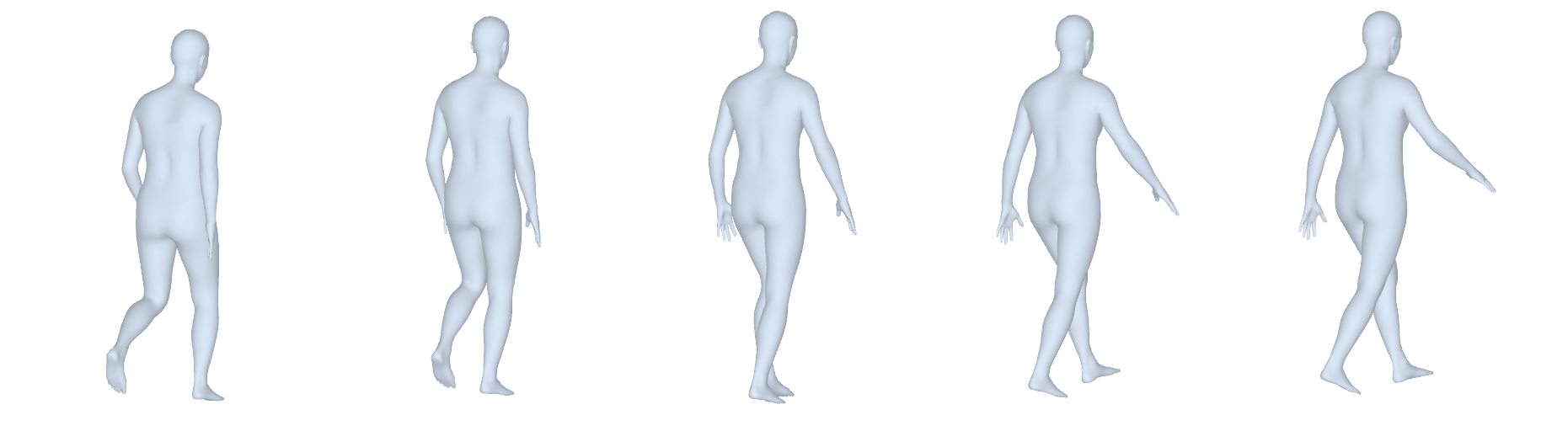}
    \hfill
    \includegraphics[width=0.49\columnwidth]{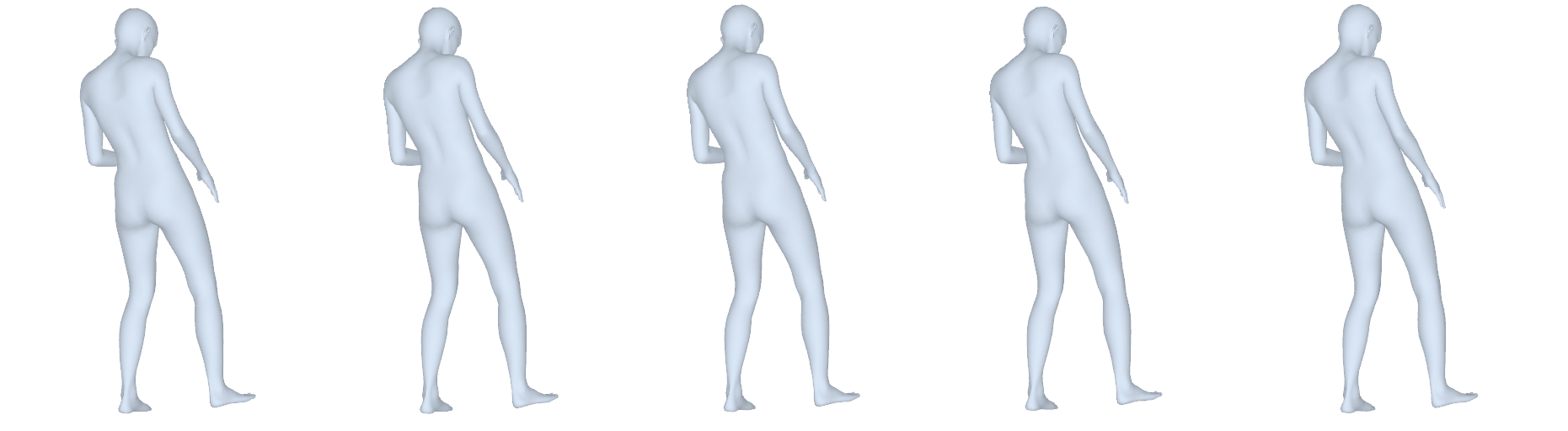}
    \hfill
    \includegraphics[width=0.49\columnwidth]{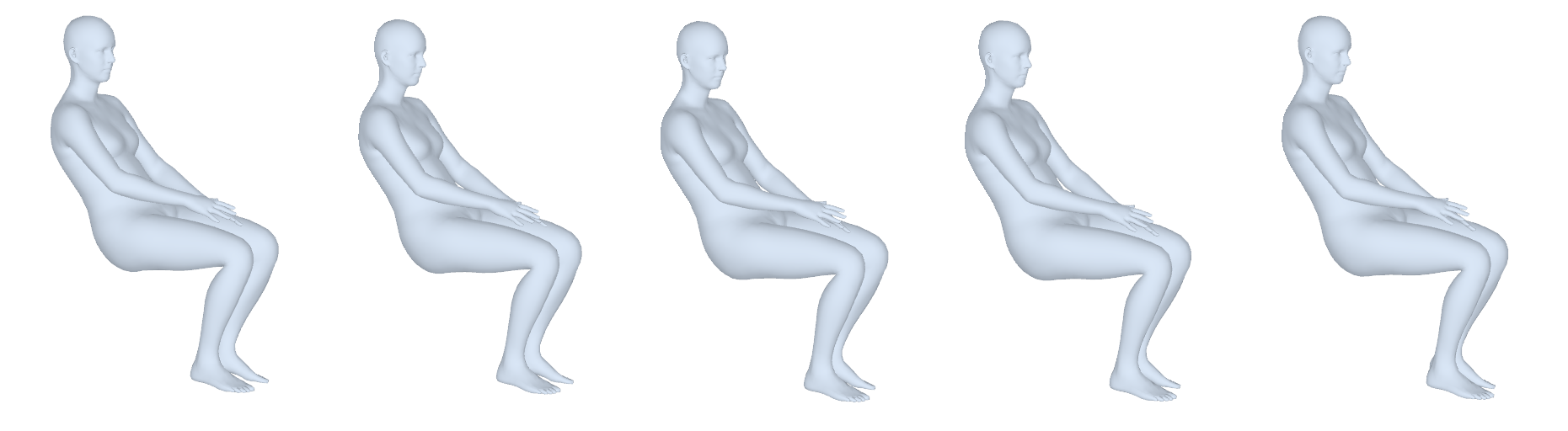}
\caption{Columns show Top-3 retrieved scenes on the test dataset for the queries “A person is walking on the crosswalk,” “A person is walking on the sidewalk,” “A person is standing on the sidewalk,” and “A person is sitting on the pavement.”}
\label{fig:motion-walking-stacked}
\end{figure*}

In our experimental setup, we use a vector database to retrieve motion videos from text queries. In Figure \ref{fig:motion-walking-stacked}, we show the three best matching scenes (by cosine similarity) for following queries: “A person is walking on the crosswalk,” “A person is walking on the sidewalk,” “A person is standing on the sidewalk,” and “A person is sitting on the pavement.” This demonstrates that the model has learned to embed the semantically salient characteristics of both motion and video modalities, enabling accurate retrieval based on nuanced differences in natural language queries.

\begin{figure}[htb]
  \centering
  \includegraphics[width=0.49\columnwidth]{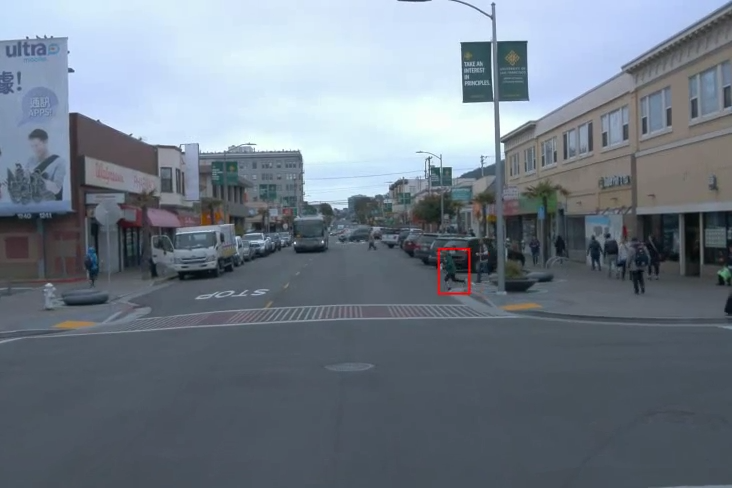}
  \hfill
  \includegraphics[width=0.49\columnwidth]{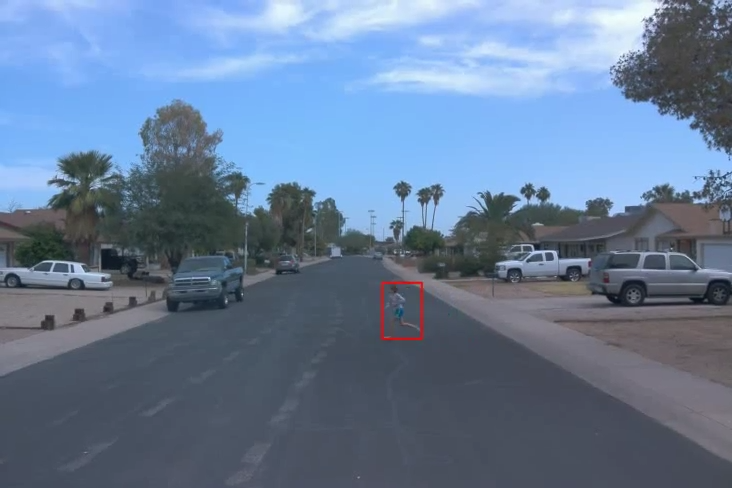}
  \\[0.5ex]
    \includegraphics[width=0.49\columnwidth]{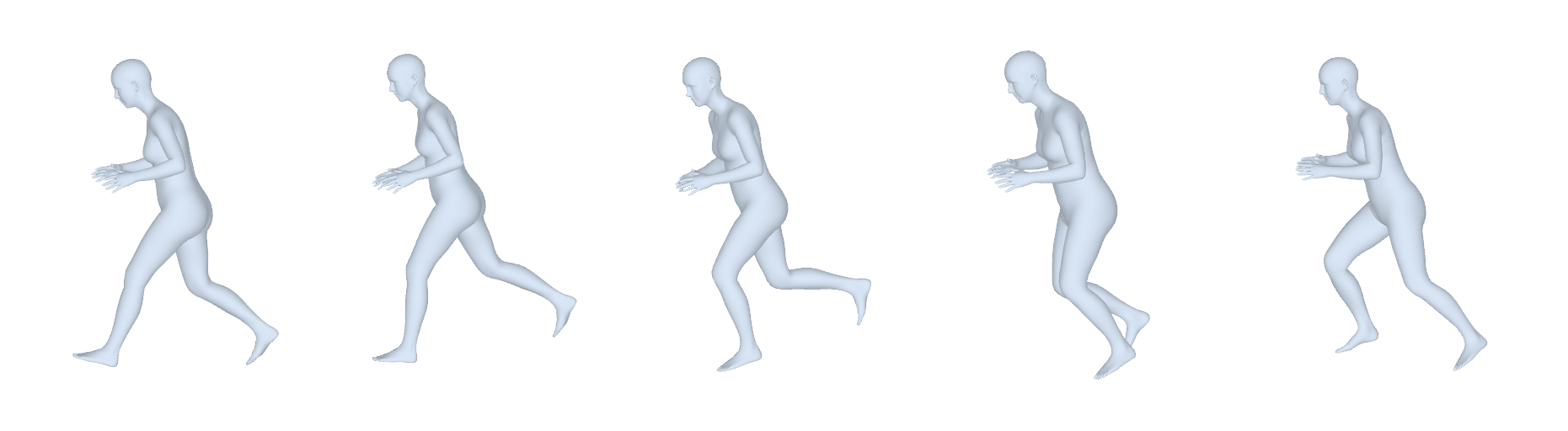}
  \hfill
  \includegraphics[width=0.49\columnwidth]{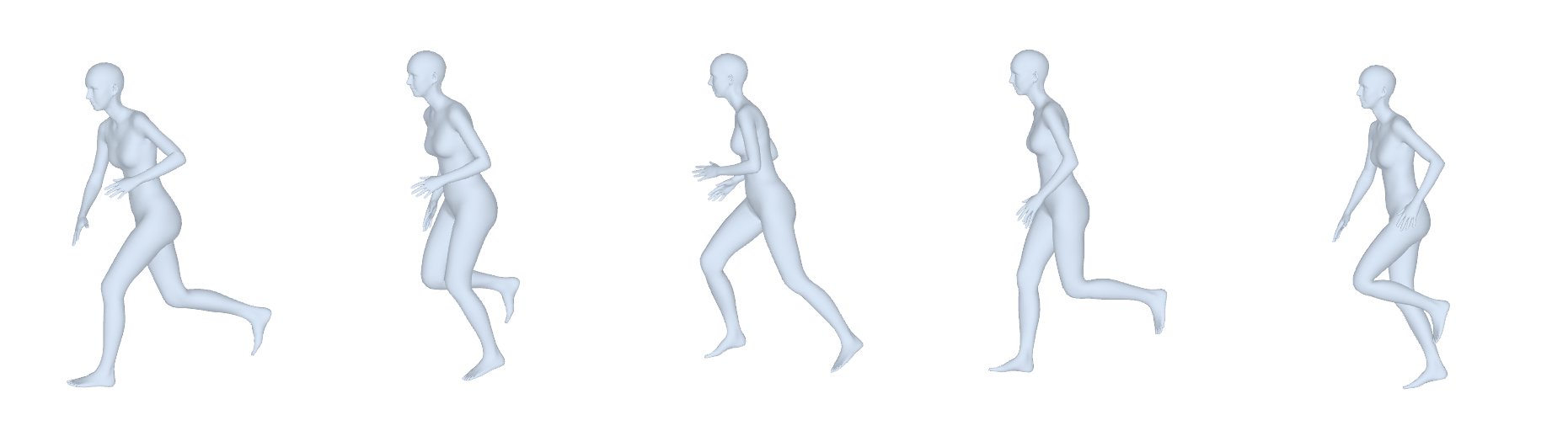}
  \caption{Retrieval of safety-critical scenes for the query “A person is running on the road.”}
  \label{fig:retrieval-saftey}
\end{figure}

Beyond showcasing our model’s ability to retrieve urban scenes involving VRUs based on context and motion, Figure~\ref{fig:retrieval-saftey} presents two safety-critical examples from the top-5 retrieval results for the query “A person is running on the road,” while the remaining non-critical results are not shown. Both scenes depict individuals suddenly running into the street ahead of the vehicle.
Since such rare events are buried in the long tail of large-scale datasets, they are difficult to find through manual inspection or random sampling. The example highlights ContextMotionCLIP's effectiveness in retrieving rare, high-risk interactions relevant to autonomous driving systems. 

\subsection{Quantitative Results}
We compare our model ContextMotionCLIP against TC-ClIP \cite{kim2024tcclip} and Vi-Fi-CLIP \cite{hanoonavificlip}. The results are reported in Table \ref{tab:topk-acc}. Our best model uses bilinear pooling to fuse motion and video features, achieving superior performance and outperforming the TC-CLIP baseline.

Compared to the baseline methods, our model achieves best accuracies for top-{1,2,3,5} with a top-1 accuracy of 48.2\% and top-5 accuracy of 69.9\%. We ouperform TC-CLIP by 27.5\% on top-1 accuracy. 
These results indicate that the strength of our approach stems from the combination of our model design and the integration of both motion and context information.

\begin{table}[htb]
  \caption{Comparison of Top-K accuracies across our model and baseline models trained with their best training method on the WayMoCo test set. }
  \centering
    \begin{tabular}{lcccc}
    \toprule
    &  \multicolumn{4}{c}{Top-K Acc (\%) $\uparrow$} \\
    \cmidrule(lr){2-5}
      Model  & K=1 & K=2 & K=3 & K=5 \\
      \midrule
      ViFi-CLIP   \cite{hanoonavificlip}             & 35.9           & 50.1          & 59.5    & 69.3 \\
      TC-CLIP \cite{kim2024tcclip}                 &     37.8       & 51.9          & 60.9       & 69.8 \\
      ContextMotionCLIP & \textbf{48.2}  & \textbf{58.3} & \textbf{62.6} &  \textbf{69.9}  \\ 
      \bottomrule
    \end{tabular}
  
  \label{tab:topk-acc}
\end{table}

\subsection{Ablation Studies}
The following ablation studies identify the primary factors contributing to our model's performance.

First, we train and evaluate our MotionCLIP model on the AMASS dataset, using the same data split as in the original MotionCLIP experiments.
Employing our specific loss function and video embedding strategy, we achieve an improvement by 14.9\% in accuracy over the original MotionCLIP, as shown in Table \ref{tab:motionclip-comparison}, after training for 540 epochs. 
We also retrain MotionCLIP to operate at 10~fps with 20‑frame input sequences, down from the original 30~fps with 60‑frame sequences. This reduces temporal resolution leads to a slight drop in accuracy compared to our 30~fps/60‑frame model—likely because the model has less temporal context to capture fine‑grained motion dynamics.

\begin{table}[htb]
  \caption{Top-1 and Top-5 accuracy comparison between MotionCLIP and our method across different frame rates.}
  \centering
  \begin{tabular}{l c c c c}
  \toprule
  & & &  \multicolumn{2}{c}{Top-K Acc (\%) $\uparrow$} \\
    \cmidrule(lr){4-5}
    Model & FPS & Sequence Length & \quad K=1 & K=5 \\
    \midrule
    MotionCLIP \cite{tevet2022motionclip} & 30 & 60 & \quad 40.9 & 57.7 \\
    MotionCLIP (ours)                     & 10 & 20 & \quad 43.4 & 58.4 \\
    MotionCLIP (ours)                     & 30 & 60 & \quad \textbf{47.0} & \textbf{64.6} \\
    \bottomrule
  \end{tabular}
  \label{tab:motionclip-comparison}
\end{table}

We further analyze the impact of red bounding boxes around the person of interest on the retrieval performance of our base model and the TC-CLIP baseline. By explicitly guiding the attention of the model to specific regions in the video leads to substantial gains across all metrics (see Table~\ref{tab:abl-bb}). Both TC-CLIP and our model significantly benefit from this simple focus mechanism, showing marked improvements in Top-K accuracy.

\begin{table}[htb]
  \caption{Comparison of Top-K accuracies (with/without bounding boxes) on the test set of our new dataset.}
  \centering
  \begin{tabular}{l c c c c c}
    \toprule
    &  & \multicolumn{4}{c}{Top-K Acc (\%) $\uparrow$} \\
    \cmidrule(lr){3-6}
    Model                     & Bounding Boxes & K=1   & K=2   & K=3   & K=5   \\
    \midrule
    TC-CLIP \cite{kim2024tcclip}             & $\times$    & 32.4 & 46.8 & 55.5 & 65.9  \\
                                            & \checkmark  & 37.8  & 51.9  & 60.9 & 69.8 \\
    ContextMotionCLIP                       & $\times$    & 43.8 & 53.6 & 58.9 & 66.1 \\ 
                                            & \checkmark  & \textbf{48.2} & \textbf{58.3} & \textbf{62.6} & \textbf{69.9} \\
    \bottomrule
  \end{tabular}
  \label{tab:abl-bb}
\end{table}
WayMoCo provides labels for both standalone motion and context categories, as well as full-sentence descriptions. We train our base model on each type and also test swapping the order of context and motion. As Table~\ref{tab:annotation-types-acc} shows, accuracies are similar across these setups, so either annotation style can be used depending on the application. Only the full-sentence annotations show a slight decrease in accuracy, suggesting that this annotation style may present a slightly more challenging learning task.
\begin{table}[htb]
  \caption{Top-K accuracies for ContextMotionCLIP on three different annotation types, each representing a distinct dataset.}
  \centering
  \begin{tabular}{l c c c c}
    \toprule
    & \multicolumn{4}{c}{Top-K Acc (\%) $\uparrow$} \\
    \cmidrule(lr){2-5}
    Annotation Type & K=1 & K=2 & K=3 & K=5 \\
    \midrule
    Full Sentences   & 47.4 & 56.7 & 61.5 & 66.6 \\
    Motion Context   & 48.2 & 58.3 & 62.6 & 69.9 \\
    Context Motion   & 48.1 & 57.6 & 62.7 & 69.5 \\
    \bottomrule
  \end{tabular}
  \label{tab:annotation-types-acc}
\end{table}

There are various loss functions for contrastive learning. In our base model, we use cosine similarity, and additionally experiment with SoftTargetCrossEntropy and InfoNCE. We also explore different fusion methods, including simple concatenation of the modality embeddings, applying self-attention to the concatenated vector, and using bilinear pooling between the modalities. Table~\ref{tab:loss-acc} provides an overview: using InfoNCE results in lower accuracy, while both SoftTargetCrossEntropy and cosine loss yield similarly strong performance. Among the fusion methods, bilinear pooling and self-attention offer slight improvements over the baseline, with bilinear pooling producing our best overall results.
\begin{table}[htb]
\caption{Top-K accuracies on the test set of our dataset for different loss functions and fusion methods used in ContextMotionCLIP.}
  \centering
  \begin{tabular}{l c c c c c}
    \toprule
    & & \multicolumn{4}{c}{Top-K Acc (\%) $\uparrow$} \\
    \cmidrule(lr){3-6}
    Loss & Fusion & K=1 & K=2 & K=3 & K=5 \\
    \midrule
    InfoNCE                  & Concat       & 42.2 & 50.5 & 55.4 & 60.8 \\
    SoftTargetCrossEntropy   & Concat       & 46.8 & 56.4 & 61.1 & 68.0 \\
    Cosine                   & Concat       & 47.6 & 56.7 & 61.2 & 66.6 \\
    SoftTargetCrossEntropy   & Self-Atten.  & 46.4 & 57.1 & 61.4 & 67.8 \\
    Cosine                   & Self-Atten.  & 46.6 & 57.1 & 62.0 & 68.5 \\ 
    SoftTargetCrossEntropy   & Bilinear    & 48.1 & 57.9 & \textbf{62.9} & 69.8 \\
    Cosine                   & Bilinear    & \textbf{48.2} & \textbf{58.3} & 62.6 & \textbf{69.9} \\   
    \bottomrule
  \end{tabular}
  \label{tab:loss-acc}
\end{table}

\section{Discussion}
Our experiments show that the proposed method outperforms state-of-the-art video retrieval models on our WayMoCo dataset, achieving a strong improvement in top-1 accuracy and a moderate gain in top-5. We attribute the smaller top-5 margin to overfitting during extended training, which could be addressed through better regularization or early stopping.

The motion annotation pipeline yields reasonable labels, but frequent errors—such as overuse of "turn"—highlight limitations of 2D pose, especially when vehicle motion around a stationary person creates misleading apparent movement. Stronger 3D pose estimates and the inclusion of 3D trajectories could enable the use of models like TMR for substantial improvement in sequence-based motion retrieval. 
Because lower-body occlusion makes pose estimation challenging for TokenHMR, it often estimates sitting poses. As a result, MotionCLIP frequently misclassifies these cases, leading to incorrect labels in the WayMoCo dataset. In contrast, our ContextMotionCLIP is often able to retrieve the correct motion sequences by leveraging contextual information, as shown in Figure~\ref{fig:motion-walking-stacked}.

CLIP is known to struggle with modeling spatial relationships between objects \cite{kamath2023whats}. This issue persists in our setup, preventing the model from effectively learning spatial relations. As a result, our queries can only detect whether an object is near a person, without capturing more detailed spatial configurations. Integrating spatially-aware pretraining as in SpatialVLM \cite{10658310}, which aligns 3D spatial relationships with language, or adopting 3D-aware transformer architectures as in SpatialCLIP \cite{Wang_2025_CVPR} could substantially improve spatial relationship modeling.

While context annotations are generally accurate, they operate at the frame level and fail to capture temporal transitions, e.g. a person stepping from a sidewalk onto the street. Additionally, the current setup relies on a fixed vocabulary. Unlike other open-vocabulary models, we observed limited generalization to out-of-dataset queries. However, when trained on full-sentence annotations, the model remains robust to variations in phrasing, allowing us to change or remove certain words from a query while still retrieving the correct scenes. Using all 600,000 different annotations during training and augmenting annotations with LLM-generated paraphrases may help improve open-vocabulary retrieval in future work.

\section{Conclusion}
We proposed a novel context-based motion retrieval framework via natural language. Our proposed method, ContextMotionCLIP, fuses SMPL-based motion representations with video data, embedding them into a shared embedding space with natural language, enabling accurate open-vocabulary retrieval of human motions and their context. We developed the WayMoCo dataset for combining motion and context and evaluated our method against the state-of-the-art. Our method achieves a 27.5\% absolute gain in top-1 retrieval accuracy over the best prior video model, with a consistent improvement in top-5 performance as well. Future work will focus on integrating higher-fidelity 3D poses and enhancing our annotations. Combined with the public release of our code, dataset, and interactive tool, these advancements will further empower targeted evaluation of autonomous driving systems in rare, safety-critical scenarios.

\bibliographystyle{IEEEtran}
\bibliography{IEEEabrv,bib}

\section*{Biography Section}

\begin{IEEEbiography}[{\includegraphics[width=1in,height=1.25in,clip,keepaspectratio]{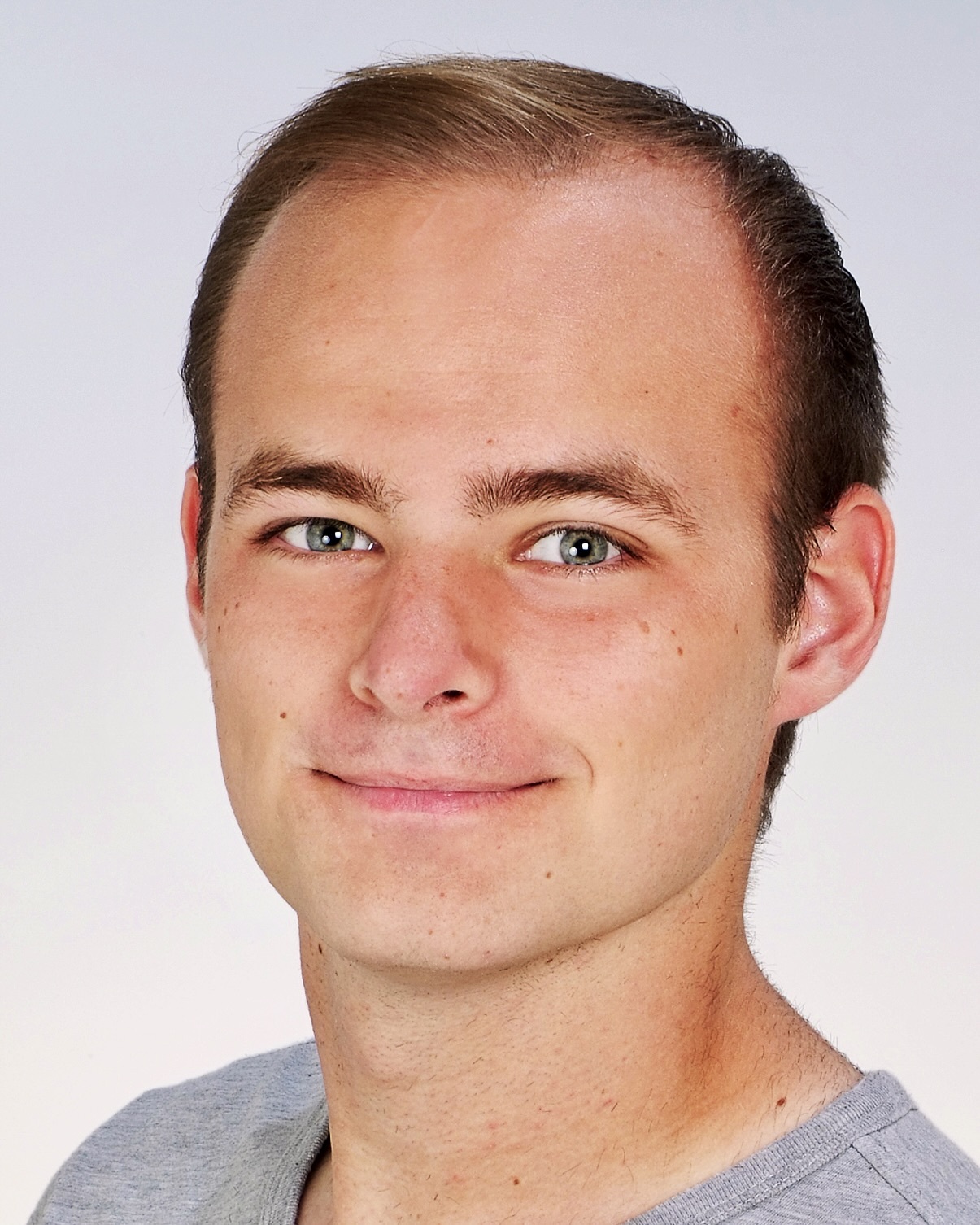}}]{Stefan Englmeier} is pursuing his Ph.D. at Munich University of Applied Sciences in the Intelligent Vehicles Lab. He earned his M.Sc. in Computer Science with a focus on Visual Computing and Machine Learning at Munich University of Applied Sciences. His research focuses on autonomous driving, specifically on trajectory planning using large language models to improve decision-making and robustness in complex driving environments.
\end{IEEEbiography}

\vspace{11pt}

\begin{IEEEbiography}[{\includegraphics[width=1in,height=1.25in,clip,keepaspectratio]{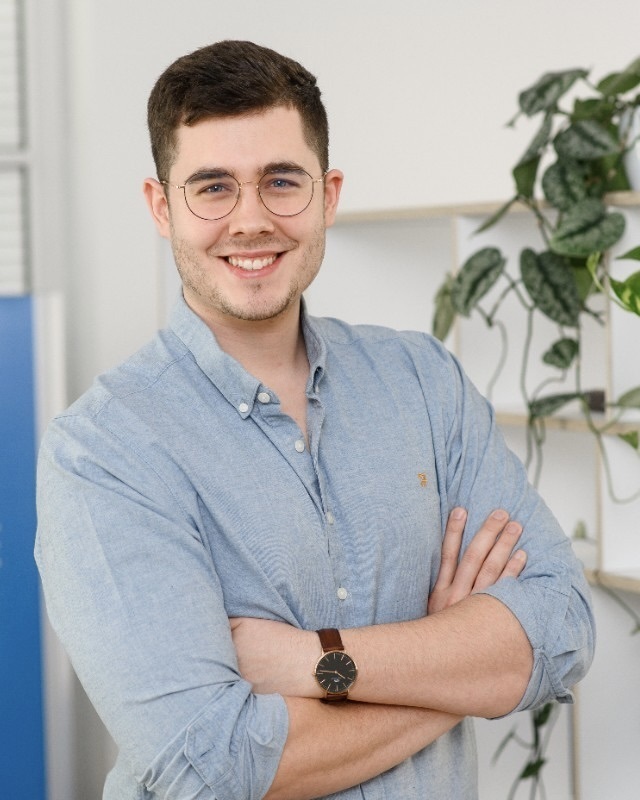}}]{Max A. Büttner}
is pursuing his Ph.D. at Munich University of Applied Sciences in the Intelligent Vehicles Lab. He earned his M.Sc. in Electrical Engineering at Munich University of Applied Sciences. His research interests focus on multimodal perception and pose estimation of VRUs for autonomous driving.
\end{IEEEbiography}

\vspace{11pt}

\begin{IEEEbiography}[{\includegraphics[width=1in,height=1.25in,clip,keepaspectratio]{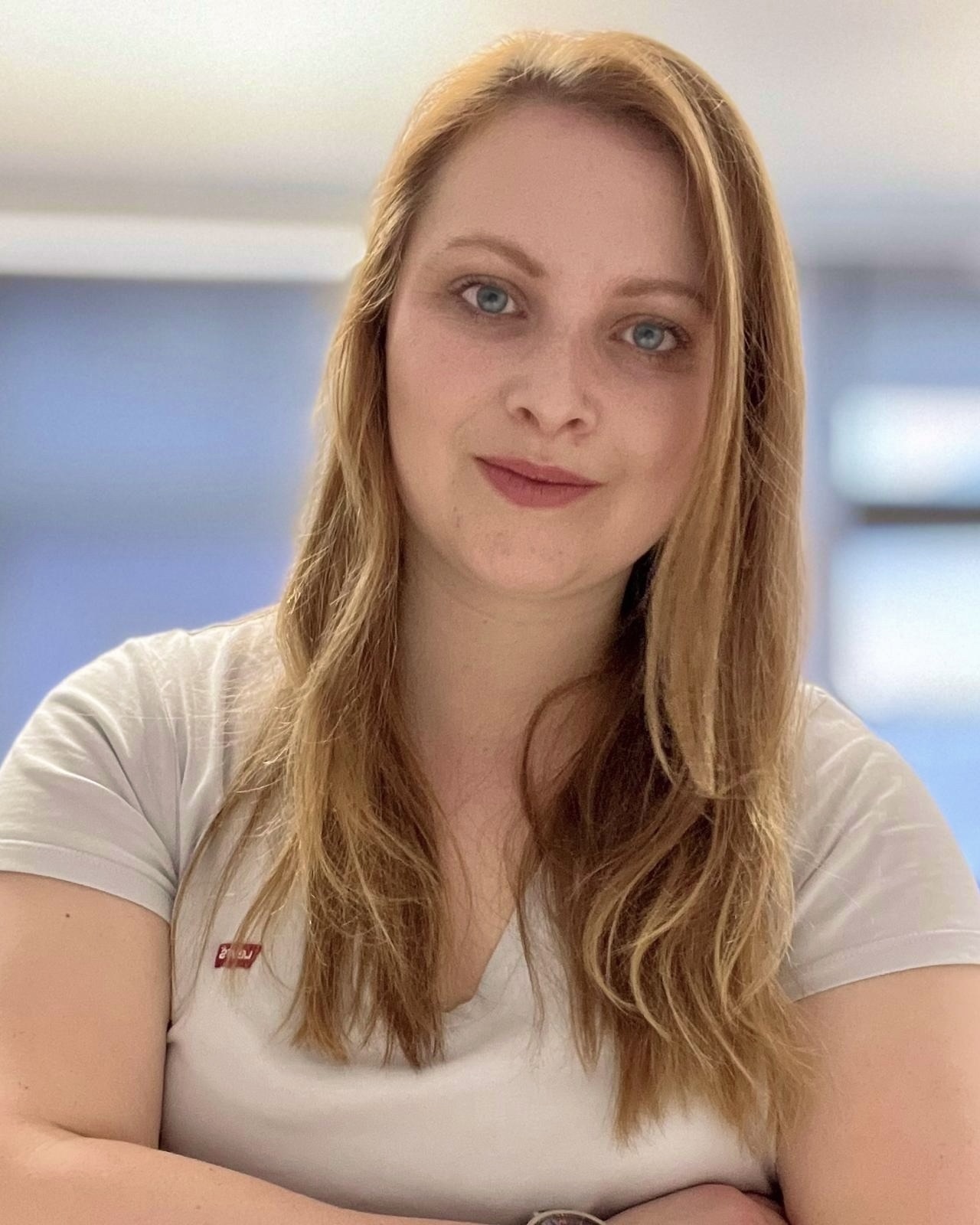}}]{Katharina Winter}
is pursuing her Ph.D. at Munich University of Applied Sciences in the Intelligent Vehicles Lab. She earned her M.Sc. in Media Informatics at LMU Munich in 2023. Her research interests include Multimodal LLMs for autonomous driving, end-to-end trajectory planning and AI explainability.
\end{IEEEbiography}

\vspace{11pt}

\begin{IEEEbiography}[{\includegraphics[width=1in,height=1.25in,clip,keepaspectratio]{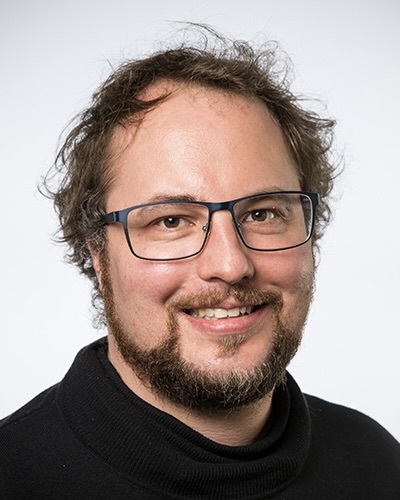}}]{Fabian B. Flohr} received his Ph.D. degree in Computer Science from University of Amsterdam, The Netherlands, in 2018. From 2012 to 2022, he was with Mercedes-Benz Research and Development in Stuttgart, Germany, where he focused on automated driving. During this time, he served as Technical Manager for Vulnerable Road User (VRU) Protection, shaping and coordinating the strategic and technical direction of the topic across multiple international teams. Since 2022, he has been a Full Professor of Machine Learning at Munich University of Applied Sciences, where he leads the Intelligent Vehicles Lab.
\end{IEEEbiography}

\vfill

\end{document}